%% file: main.tex
\documentclass{article}

\usepackage{microtype}
\usepackage{graphicx}
\usepackage{booktabs} 
\usepackage{caption}
\usepackage{subcaption}

\usepackage{multirow}
\usepackage{float}
\usepackage{placeins}

\usepackage{amsmath}
\usepackage{amssymb}
\usepackage{tcolorbox}
\tcbset{boxsep=0mm,boxrule=0pt,colframe=white,arc=0mm,left=0.5mm,right=0.5mm}
\usepackage{hyperref}
\usepackage[normalem]{ulem}

\usepackage[accepted]{icml2021}

\icmltitlerunning{This Looks Like That... Does it?  Shortcomings of Latent Space Prototype Interpretability in Deep Networks}

\begin{document}

\twocolumn[
\icmltitle{This Looks Like That... Does it? \\
          Shortcomings of Latent Space Prototype Interpretability in Deep Networks}



\icmlsetsymbol{equal}{*}

\begin{icmlauthorlist}
\icmlauthor{Adrian Hoffmann}{equal,ethI}
\icmlauthor{Claudio Fanconi}{equal,ethE}
\icmlauthor{Rahul Rade}{equal,ethE}
\icmlauthor{Jonas Kohler}{ethI}
\end{icmlauthorlist}

\icmlaffiliation{ethI}{Department of Computer Science, ETH Zurich, Switzerland}
\icmlaffiliation{ethE}{Department of Information Technology and Electrical Engineering, ETH Zurich, Switzerland}

\icmlcorrespondingauthor{Adrian Hoffmann}{adriahof@ethz.ch}

\icmlkeywords{Machine Learning, ICML}

\vskip 0.3in
]



\printAffiliationsAndNotice{\icmlEqualContribution} 

\begin{abstract}
Deep neural networks that yield human interpretable decisions by architectural design have become an increasingly popular alternative to post hoc interpretation of traditional black-box models. Among these networks, the arguably most widespread approach is so-called prototype learning, where similarities to learned latent prototypes serve as the basis of classifying unseen data. In this work, we point to a crucial shortcoming of such approaches. Namely, there is a semantic gap between similarity in latent space and input space, which can corrupt interpretability. We design two experiments that exemplify this issue on the so-called ProtoPNet. We find that its interpretability mechanism can be led astray by crafted noise or JPEG compression artefacts, which can lead to incoherent decisions. We argue that practitioners ought to have this shortcoming in mind when deploying prototype-based models in practice.

\end{abstract}

\input{intro}

\input{background}


\input{head_on_stomach}

\input{jpeg_noise}

\input{discussion}

\input{conclusion}

\bibliography{main}
\bibliographystyle{icml2021}


\input{appendix}

\end{document}

%% file: intro.tex

\section{Introduction}
Due to the increasingly widespread application of artificial neural networks in science, industry, and society, interpreting decisions made by these models is becoming more and more important \cite{samek2019towards, rudin2019stop}. While many works in this regard have focused on ex-post interpretation of black-box machine learning models (e.g. \citet{bach2015pixel,kim2018interpretability,bau2017network,lapuschkin2019unmasking}), a recent line of research breaks with this paradigm by proposing to develop network architectures that are inherently interpretable by design (e.g. \citet{girdhar2017attentional,chen2018looks,li2018deep,brendel2019approximating}). These models aim at basing their decisions on a small number of human-understandable features that allow users to comprehend how predictions come about while keeping compromises in terms of predictive performance low \cite{chen2018looks}.

In this work, we focus on prototype-based reasoning which labels datapoints based on distances to learned prototypes of each class in latent space \cite{li2018deep,chen2018looks,gee2019explaining,ming2019interpretable,xu2020attribute}. While these methods undoubtedly yield interpretable decisions in many cases, we point to a key limitation of such approaches: 

\begin{tcolorbox}
 The fact that two datapoints have a similar latent representation does not necessarily entail that they share similarities in terms of human-interpretable features in the input space. 
\end{tcolorbox}
\vspace{-1mm}

In the remainder of this paper, we substantiate this claim by closely investigating the performance of an image classification network termed \textit{Prototypical Part Network (ProtoPNet)}. This network is proclaimed to yield a transparent and human-like decision process by classifying images based on similarity scores between (parts of) the test image and (parts of) the training images, called prototypes, in latent space~\cite{chen2018looks}. To assess these claims, we first argue that, for the reasoning of ProtoPNet to be human-interpretable, the following statement must hold.\looseness=-1

\noindent \textbf{Equivalence 1.} 
\vspace{-3mm}
\begin{center} \label{eq:equivalence}
\textit{Two image patches look similar to a ProtoPNet\\ $\Longleftrightarrow$ Two image patches look similar to a human}
\end{center}
\vspace{-2mm}


In the following, we design two experiments, which demonstrate that neither of the two directions always holds: (i) We first show that small, human-imperceptible perturbations in a given test image can make a properly trained ProtoPNet identify prototypes in arbitrary regions (for the human eye) of the input (Section~\ref{sec:head_on_stomach}). (ii) Secondly, we show that simple JPEG compression can lead to drastically lower similarity scores between image patches that look similar to a human before and after compression (Section~\ref{sec:JPEG}).

%% file: background.tex
\section{Preliminaries}
\subsection{ProtoPNets: Intuition and Inner Workings}
A ProtoPNet~\cite{chen2018looks} is comprised of three stages: (i) First, an image is fed through a convolutional backbone $f$ (e.g. ResNet \cite{resnet}) to extract meaningful latent representations. These representations are then fed through (ii) a prototype layer which computes $L^2$ norm-based similarities to learned, class-specific prototypes in the latent space and then (iii) classify the image by passing these similarity scores through a softmax classification layer.

After classification, image patches and prototypes with the highest similarity scores can be visualized in input space in order to represent the network's decision in a human-perceptible way. As a result, ProtoPNets have some level of interpretability built directly into the architecture and the authors in \citet{chen2018looks} even go so far as to attest their network a transparent, human-like reasoning process. 

\textbf{Prototype Layers.} Given an input image $\mathbf{x}$ and its latent representation $\mathbf{z}=f(\mathbf{x})$ of size $H\times W\times D$, ProtoPNet computes similarity scores between all pixels in $\mathbf{z}$ and all $m$ learnable prototypes $\mathbf{P}=\{\mathbf{p}_l\}_{l=1}^m$ of size $1\times 1 \times D$. For each pixel $\mathbf{z}_k\in\mathbb{R}^D$ and prototype $\mathbf{p}_l\in\mathbb{R}^D$, this score is:
\begin{equation} \label{eq:similarity}
s(\mathbf{z}_k,\mathbf{p}_l)=\log\left(\frac{\|\mathbf{z}_k-\mathbf{p}_l\|_2+1}{\|\mathbf{z}_k-\mathbf{p}_l\|_2+\epsilon}\right),\quad \epsilon>0. 
\end{equation}
This results in $m$ heatmaps of the same spatial size as $\mathbf{z}$ which indicate where and how strongly a given prototype is present in $\mathbf{z}$. Since spatial relations are preserved in the convolutional backbone $f$, the heatmaps can be up-sampled and visualized on the input image. For each prototype $\mathbf{p}_l$, simple max-pooling over the spatial dimensions yields the final similarity score $s_{\mathbf{p}_l}(\mathbf{z})$. Hence, the output of a prototype layer is a vector of $m$ similarity values.

Importantly, the prototypes are class-specific and learned employing a contrastive loss as a regularizer that incentivizes a semantically meaningful clustering structure in the latent space (see \citet{chen2018looks}, Section 2.2). Furthermore, as the final step of training, the prototypes $\mathbf{p}_l$ (learned via backpropagation) are projected onto the nearest neighbour $\mathbf{z}^i_k$ of all inputs $x^i$ of the prototype-specific class. This last step is crucial for the proclaimed interpretability, as every prototype is thereby being equated with the latent representation of a patch of one of the training images and hence can be visualised in a human-perceptible way.
    
\subsection{Projected Gradient Descent}
We use an adapted version of projected gradient descent (PGD) \cite{madry2019deep} for our experiments in Section~\ref{sec:head_on_stomach}. PGD is commonly used in the context of adversarial attacks \cite{akhtar2018threat} where it generates a perturbed image from a correctly classified sample. The ultimate goal is that the resulting image is misclassified by the given model while keeping the perturbation below a threshold in a given norm. We here reuse the idea behind PGD but pursue a different goal i.e. the perturbation aims to make the network produce absurd interpretations instead of misclassifying the sample.

\subsection{Compression and other artefacts}
In Section~\ref{sec:JPEG}, we investigate the robustness of ProtoPNet's interpretation mechanism against common file format artefacts. Specifically, we study JPEG compression, a lossy image compression format \cite{Penn92}. We choose these artefacts for their convenience. However, we suspect that our results, which indicate that ProtoPNets can fail to convey their reasoning, also apply to other noises (like Rayleigh in MRI or speckle in ultrasound imaging \cite{dietrichMRI2008, goyalBiomedPharmaJournal2018}). While it has been noted before that neural networks are not robust to compression artefacts in classification settings \cite{hendrycksICLR2019}, we are unaware of any prior results on their impacts on prototype-based interpretability mechanisms.

\begin{figure*}[th]
\centering
\begin{subfigure}{0.32\textwidth}
  \centering
  \includegraphics[width=\linewidth]{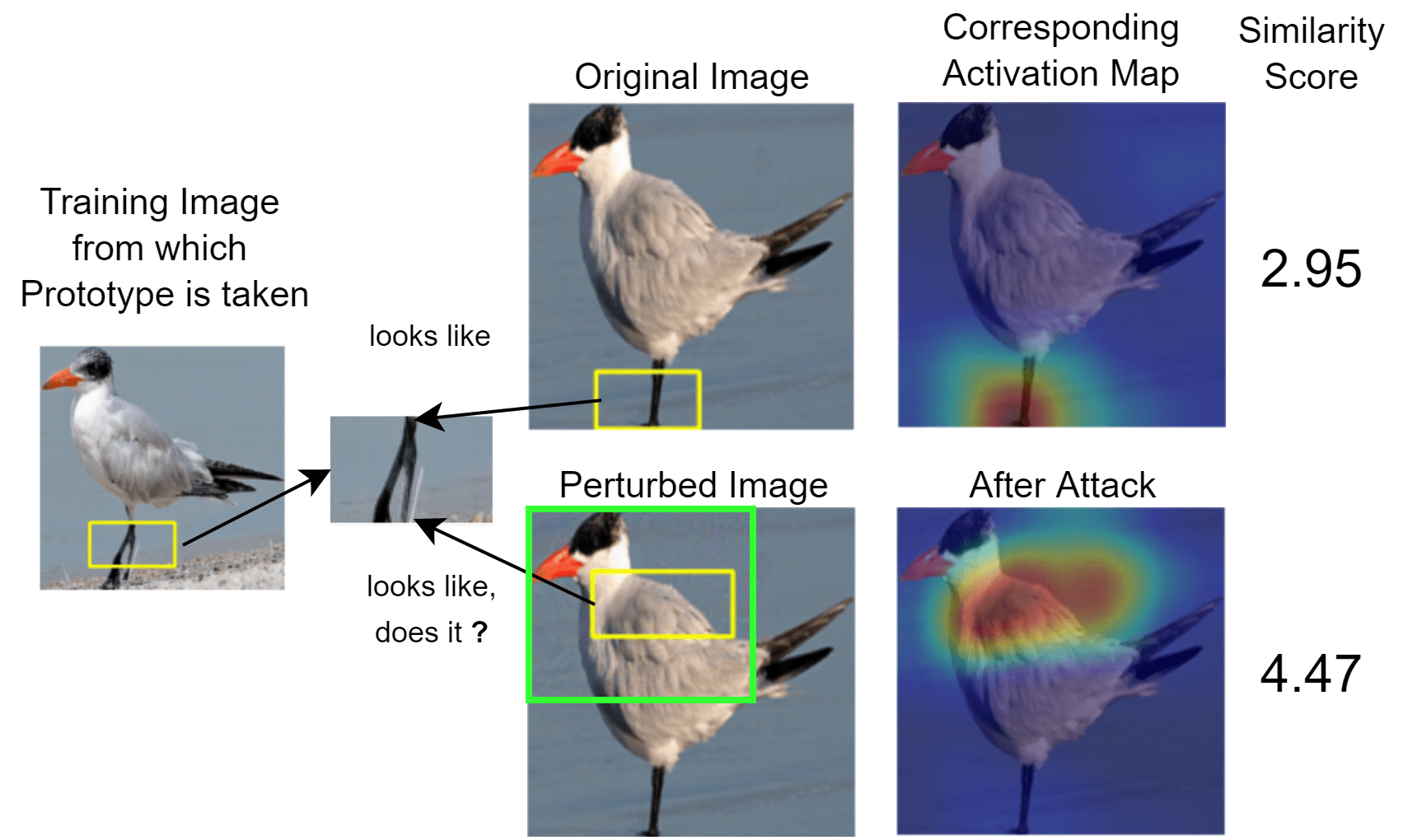}
  \caption{ResNet-18}
  \label{fig:results_adv_1}
\end{subfigure}\hfill
\begin{subfigure}{0.32\textwidth}
  \centering
  \includegraphics[width=\linewidth]{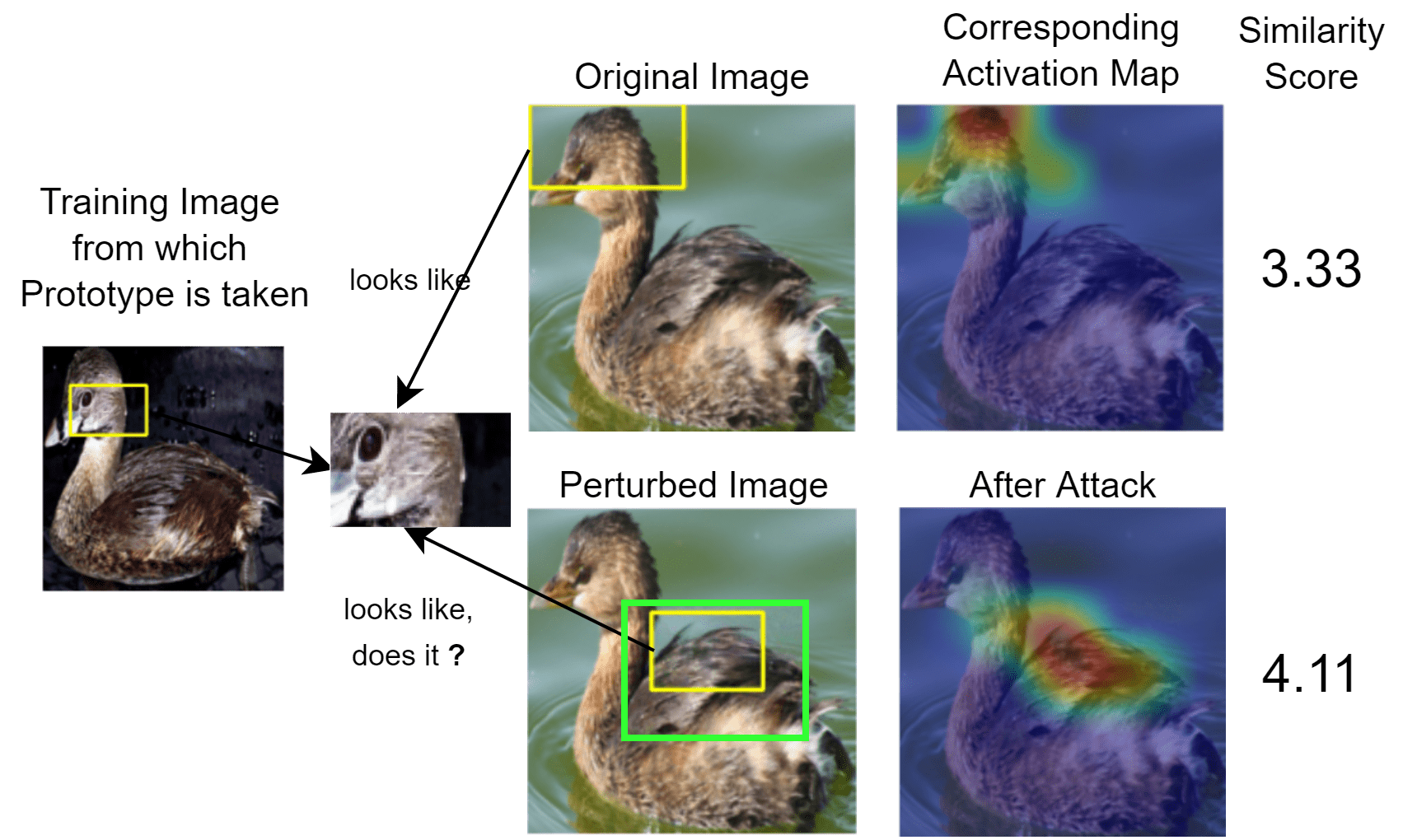}
  \caption{ResNet-34}
  \label{fig:results_adv_2}
\end{subfigure}\hfill
\begin{subfigure}{0.32\textwidth}
  \centering
  \includegraphics[width=\linewidth]{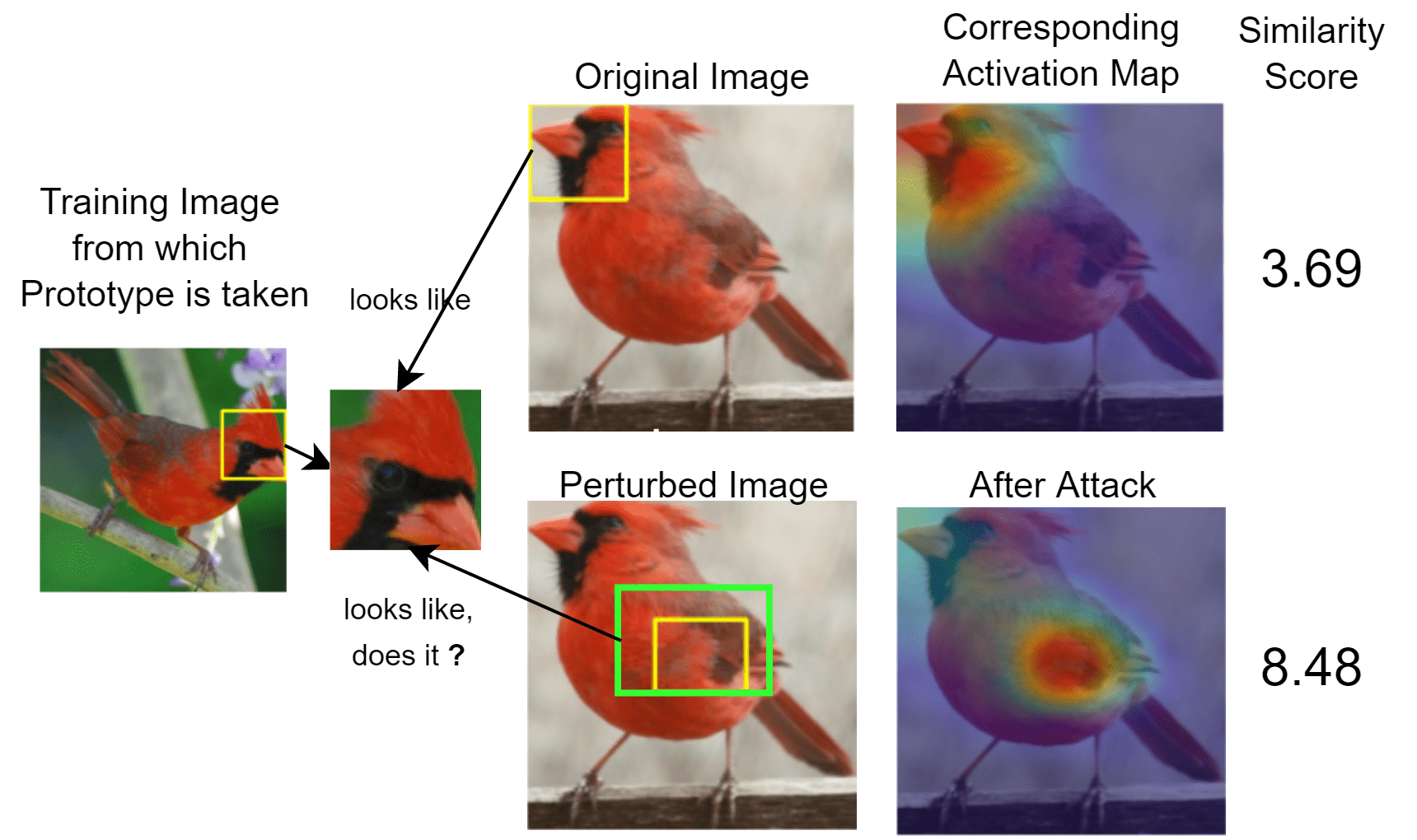}
  \caption{VGG-19}
  \label{fig:results_adv_3}
\end{subfigure}\hfill
\vspace{-1mm}
\caption{Three examples (one per backbone) for the Head on Stomach experiment. The first image is the source of the prototype we consider in the example. The next image shows the prototype itself. The top row of the remaining four images shows where the respective ProtoPNet finds the shown prototype in the input image (yellow rectangle and heat map). The bottom row shows where it recognises the prototype after we add the noise. The position of the added noise is shown by the green rectangle.}
\label{fig:results_adv}
\end{figure*}

\subsection{Definition of Interpretability}
Lastly, we define the notion of interpretability used in this work, since we are unaware of a universally agreed-upon definition. The following taken from \citet{GilpinIEEE2018} suits our understanding: "\textit{for a system to be interpretable, it must produce descriptions that are simple enough for a person to understand using a vocabulary that is meaningful to the user.}" Additionally, we require that the description also has to be credible. A description can be simple and understandable, but if it does not match the model's internal reasoning there is no value in it. We also believe that this definition mirrors the exposition that \citet{chen2018looks} had in mind when arguing that ProtoPNet was human-interpretable. In App.~\ref{app:deriving_eq}, we show how this definition yields Equivalence~\ref{eq:equivalence}.



%% file: head_on_stomach.tex
\section{Head on Stomach Experiment}\label{sec:head_on_stomach}

In this section, we show that the left-to-right implication in Equivalence~\ref{eq:equivalence} does not always hold. Furthermore, we restrict the evaluation to test images that are correctly classified (before and after the experiment).\\
The basic idea is to take a correctly classified test image $\mathbf{x}$, choose one of the most activated prototypes the ProtoPNet identifies in a sensible location in $\mathbf{x}$, say $\mathbf{p}_l$, and then perturb $\mathbf{x}$ such that the network finds $\mathbf{p}_l$ with high similarity in a nonsensical place. All the while keeping the perturbation noise very small. An example is depicted in Figure~\ref{fig:results_adv_3} where a ProtoPNet accurately reports a similarity between the bird's head and a prototype depicting the head of a bird of the same species. However, after adding human-imperceptible noise the network reports an even higher similarity to the bird's stomach (hence the name of this experiment).

\textbf{Methodology: }For a single run of this experiment, we consider a ProtoPNet with convolutional backbone $f$, a correctly classified test image $\mathbf{x}$, and a chosen prototype $\textbf{p}_l$. Furthermore, $S$ denotes the set of pixels in the latent representation of $\mathbf{x}$ with the highest similarities to $\textbf{p}_l$ and $S_{\text{noisy}}$ is the set of latent pixels we want to move the highest similarity to. Lastly, $g_{\textbf{p}_l}(\mathbf{z})$ denotes the similarity map between prototype $\textbf{p}_l$ and latent representation $\mathbf{z}$ where each entry in $g_{\textbf{p}_l}(\mathbf{z})$ is computed with the similarity score in Eq.~\eqref{eq:similarity}. 
To achieve our goal, we formulate the following objective:
\begin{equation}\label{eq:head-on-stomach-objective}
    \begin{aligned}
    \mathcal{L}(\mathbf{x}; S, S_{\text{noisy}}) &= \frac{1}{|S_{\text{noisy}}|}\sum_{(i, j)~\in ~S_{\text{noisy}}}{g_{\textbf{p}_l}(f(\mathbf{x}))}_{ij} \\
    &~- \frac{1}{|S|}\sum_{(i, j)~\in~S}{g_{\textbf{p}_l}(f(\mathbf{x}))}_{ij}
    \end{aligned}
\end{equation}
which encourages high similarity between the latent pixels at the positions in $S_{\text{noisy}}$ and prototype $\textbf{p}_l$ while reducing similarity between the latent pixels at positions in $S$ and $\textbf{p}_l$. This allows us to state the full optimization problem:
\begin{equation}\label{eq:head-on-stomach-optimization}
\min_{\mathbf{\delta} \in \mathbb{R}^{\text{dim}(\mathbf{x})}} -\mathcal{L}\left(\mathbf{x}+\mathbf{\delta} ; S, S_{\text{noisy}}\right) 
\quad \text { s.t. } \|\mathbf{\delta}\|_{\infty} \leqslant \varepsilon
\end{equation}
where the constraint makes sure that the noise stays small.

We would like to point out two subtle differences between the usual application of PGD \cite{madry2019deep} and our approach. First, as alluded to before, our goal is not to create an adversarial example for misclassification but we rather focus on how the similarity of the chosen prototype changes.\footnote{All noisy images were still correctly classified.} Secondly, we do not add noise to the entire image but only to those patches relevant for the similarity scores.

\textbf{Results: }In total, we trained three ProtoPNets, one for each of ResNet-18, ResNet-34, and VGG-19 \cite{Simonyan15} as backbone. For training we used the same dataset (CUB-200-2011 \cite{WahCUB_200_2011}), hyper-parameters, and data-augmentation as \citet{chen2018looks}. We use PGD with $\varepsilon = 8/255$, step size $2/255$ and 40 iterations to solve the optimization problem.\\
Figure~\ref{fig:results_adv} depicts three examples of the Head on Stomach experiment (one for each backbone). As evident, the provided interpretations seem plausible on the clean images. However, when we perturb the image, the network's reasoning becomes clearly incomprehensible to humans. Additionally, the similarity scores for the noisy pictures are remarkably high (a similarity score $\geq 4$ is empirically very high, see plots in App.~\ref{app:jpeguniform}). This shows that what a ProtoPNet finds similar need not look similar to humans. More examples can be found in App.~\ref{app:hosexamples}.

%% file: jpeg_noise.tex
\section{JPEG Experiment}\label{sec:JPEG}
\begin{figure*}[h]
\begin{subfigure}{0.32\textwidth}
  \centering
  \includegraphics[width=\linewidth]{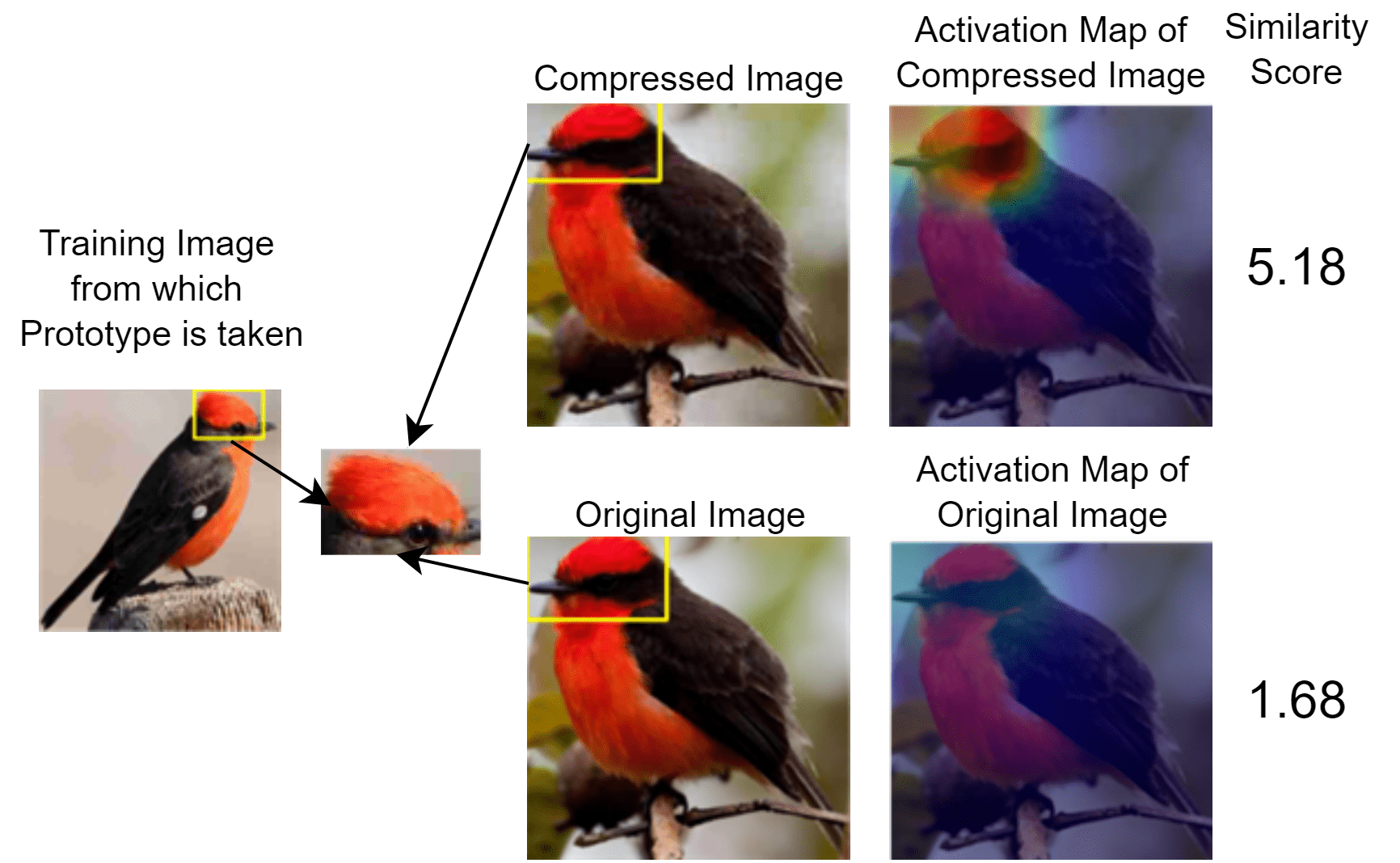}
  \caption{ResNet-18}
  \label{fig:results_jpeg_1}
\end{subfigure}\hfill
\begin{subfigure}{0.32\textwidth}
  \centering
  \includegraphics[width=\linewidth]{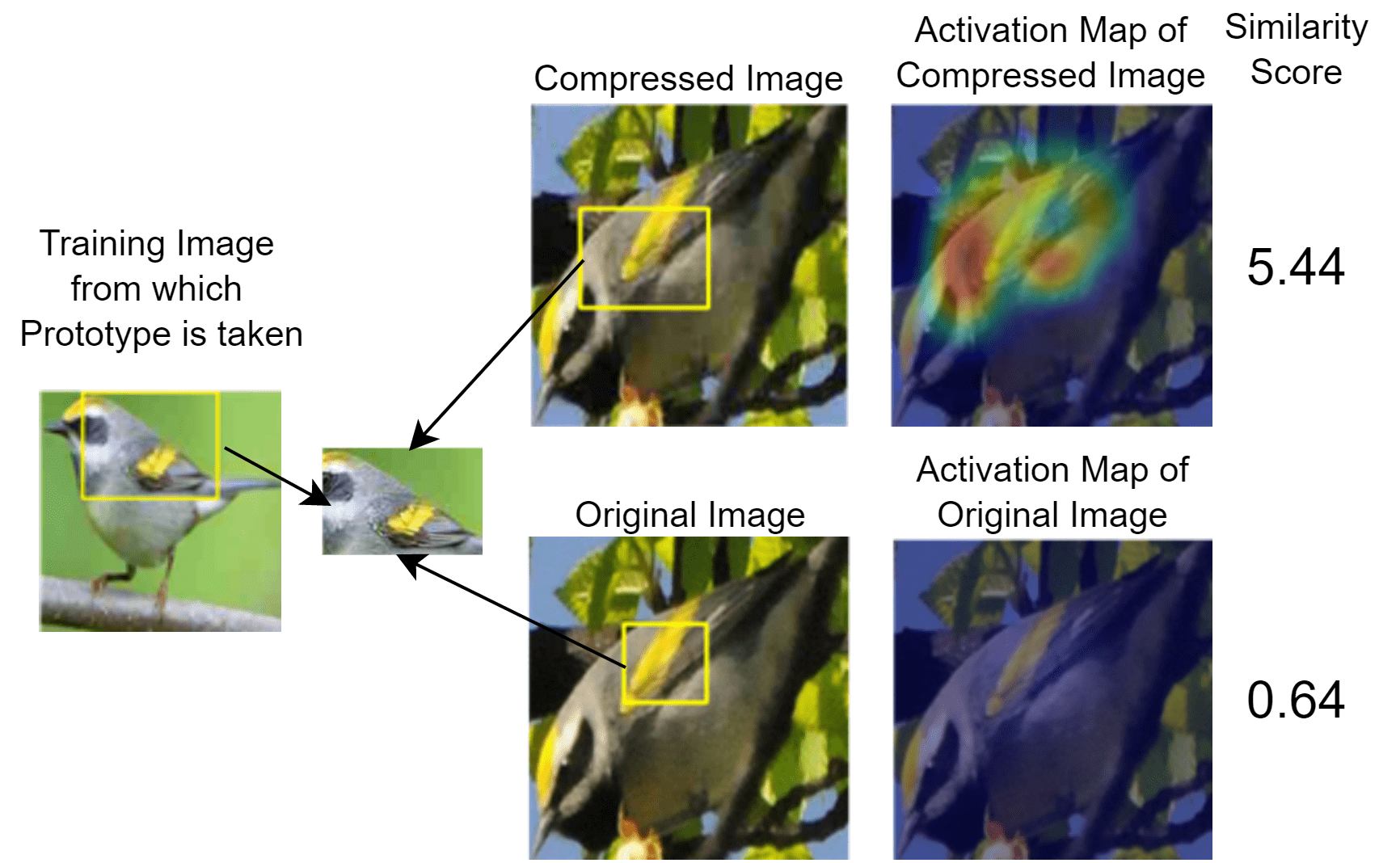}
  \caption{ResNet-34}
  \label{fig:results_jpeg_2}
\end{subfigure}\hfill
\begin{subfigure}{0.32\textwidth}
  \centering
  \includegraphics[width=\linewidth]{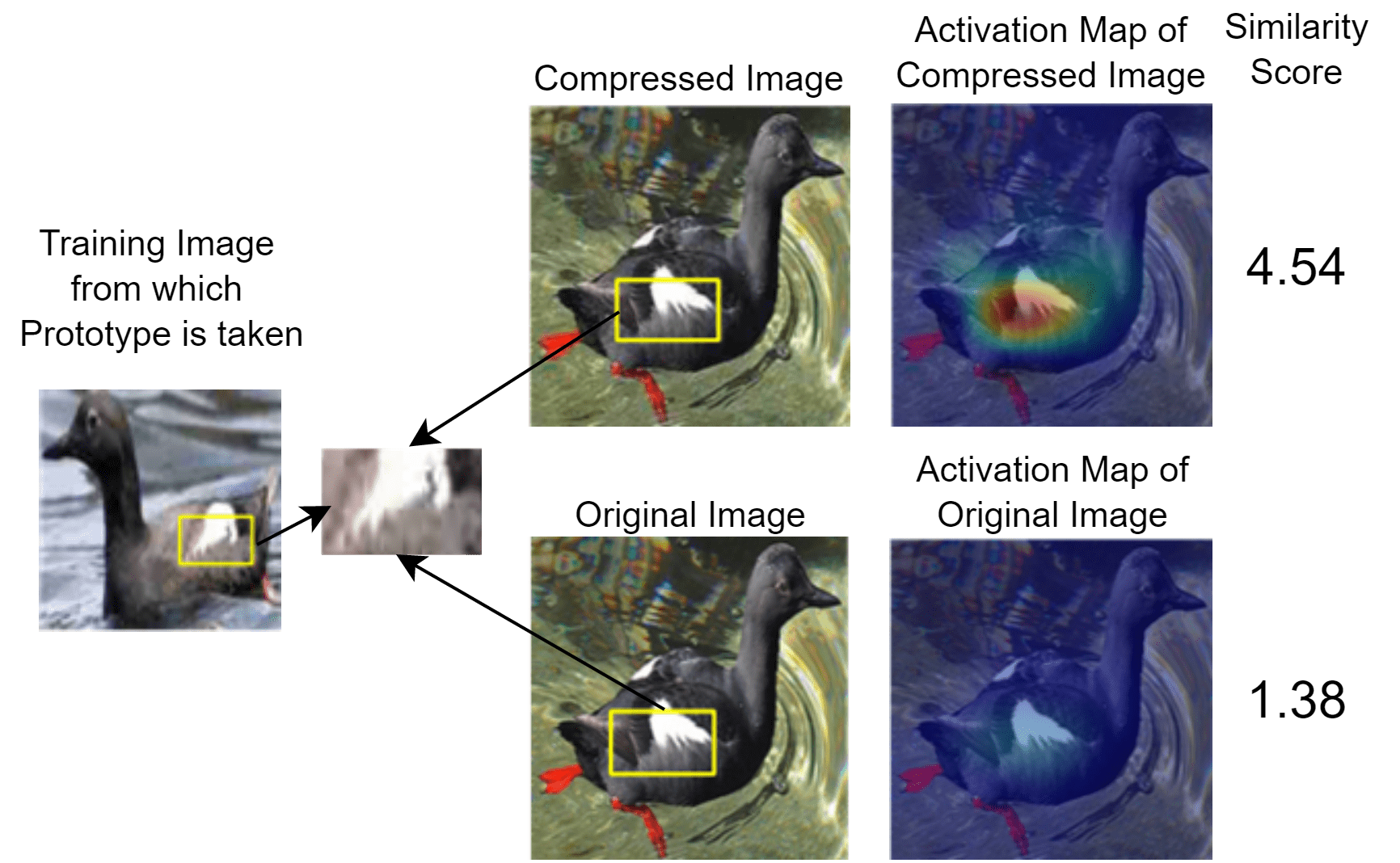}
  \caption{VGG-19}
  \label{fig:results_jpeg_3}
\end{subfigure}\hfill
\vspace{-1mm}
\caption{Three examples (one per backbone) for the JPEG experiment. In each example, the first image is the source of the considered prototype. The next image shows the prototype itself. To the right, the top (bottom) row shows a compressed (original) test image and its activation map. Similarity scores drop significantly if no JPEG compression is applied, despite the fact that the images look similar to the human eye, which highlights that ProtoPNet no longer considers the image patches similar.}
\label{fig:results_jpeg}
\end{figure*}
We next show that the right-to-left implication in Equivalence~\ref{eq:equivalence} can break down in fairly realistic scenarios. Interestingly, our experiment also uncovers a potential gap between the interpretability and decision mechanism in ProtoPNets, which is important for practitioners to keep in mind.\\
JPEG compression is a lossy compression algorithm, which leaves artefacts behind. While this noise is usually barely perceptible for humans, it may still confuse neural networks in detrimental ways \cite{hendrycksICLR2019}. In the following, we examine changes in interpretability when training a ProtoPNet on a dataset where the training images of a few (but not all) classes are JPEG compressed, which might occur e.g. when collecting training data from different sources. In particular, we assess if the interpretation provided by the model differs depending on whether the compressed or clean version of a test image is considered. As the noise is almost unrecognisable to humans, ProtoPNet would have to provide similar interpretations in both cases for the right-to-left implication in Equivalence~\ref{eq:equivalence} to hold.



\textbf{Methodology: }For this experiment, we again train on the CUB-200-2011 dataset. But first, we draw 100 classes (half of the classes) at random and apply JPEG compression with $20\%$ quality to the images of these classes (see Fig. \ref{fig:jpeg_compression}).\\
To assess whether the interpretations stay consistent for images with and without compression artefacts, we consider a given test image $\mathbf{x}$ of a compressed class and check if the compressed version of $\mathbf{x}$ (denoted $\mathbf{\bar{x}}$) is classified correctly. If so, we record the similarity score for the most activated prototype (denoted $\mathbf{p}_l$) for $\mathbf{\bar{x}}$ and track how the score of this prototype changes when we instead feed $\mathbf{x}$ into the network. If the scores for $\mathbf{p}_l$ differ significantly, then the ProtoPNet does not consider $\mathbf{x}$ and $\mathbf{\bar{x}}$ similar in the same manner as humans would, which in turn suggests that the right-to-left implication does not always hold.

\textbf{Results: }We train one ProtoPNet for each of the ResNet-18, ResNet-34, and VGG-19 convolutional backbones with the same training setup as in Section~\ref{sec:head_on_stomach}, besides the aforementioned changes to the dataset. In Figure~\ref{fig:results_jpeg}, we display three instances of the JPEG experiment. As can be seen in the top row, the ProtoPNets detect strong similarities between prototypes and image patches in accordance with human reasoning when considering compressed images. However, the similarity scores drop significantly when we remove the compression artefacts (bottom row). This suggests that (a) images that humans find very similar need not look similar to ProtoPNets and (b) the compression artefacts themselves are important for the decision making of the network\footnote{Which is  supported by a drop in accuracy (see App.~\ref{app:ppnetperformances})}. Moreover, the interpretability mechanism does not communicate the importance of these artefacts in a meaningful manner. More examples can be found in App.~\ref{app:jpegexamples}. Importantly, we note that switching from the compressed to the original test image does not simply uniformly scale down all similarity scores (see App.~\ref{app:jpeguniform}).

%% file: discussion.tex
\section{Discussion} \label{sec:discussion}

In summary, our findings suggest a gap between ProtoPNet's understanding of similarity and that of humans. While this is not severe in the case of bird images, where any flaw in interpretability is easily detected, we consider this lack of robustness to be much more critical in settings that humans are less well suited for evolutionarily, such as in medical imaging, where one might e.g. be interested in gaining insights about diseases via the neural network's decision process. Here, it is important for the network to accurately report the patterns it picks up on. While it is arguably true, that both adversarial and compression noise are known weak spots of neural networks, we want to emphasize that their application to interpretability mechanisms is novel.

Finally, we point out that we do not dispute the general usefulness of prototype-based interpretation mechanisms but rather this paper is to be understood as a call for caution. In fact, counter measures such as adversarial training and data augmentation increase the robustness of ProtoPNets albeit at the cost of reduced accuracy (see App.~\ref{sec:countermeasures}).

%% file: conclusion.tex
\section{Summary and Future Work}
In this work, we investigated potential shortcomings of the interpretability mechanism of so-called ProtoPNets, which base their decisions on similarities between image patches and visualisable prototypes. We showed that, while ProtoPNets usually bring about human-understandable decisions, there are certain cases, namely in the presence of adversarial or compression noise, where the analogy between human and network reasoning breaks down. While the first scenario is not overly surprising, the latter is of particular relevance to practitioners who might have different sources of noise in their data (e.g. lighting, device, zoom and motion artefacts). We thus caution practitioners to be mindful of the above shortcomings when using prototypical networks, especially in settings where possible flaws in the interpretability mechanism are less obvious, for example in medical imaging.

Going forward, we consider a more comprehensive study of the robustness of interpretability mechanisms to be an interesting direction of future research, including both the examination of interpretable architectures different to ProtoPNets (e.g. \citet{nakka2020robust}) as well as the identification of further sources of artefacts/noise that may lead to a break down of interpretability in real-world settings. Finally, building upon our initial results in App.~\ref{sec:countermeasures}, further improving the robustness of interpretable models via architectural enhancement as well as advanced training techniques constitutes an interesting line of future work.

%% file: appendix.tex
\newpage
\onecolumn

\appendix

\section{Deriving Equivalence~\ref{eq:equivalence}} \label{app:deriving_eq}
In this section, we motivate our claim that Equivalence~\ref{eq:equivalence} is necessary for a ProtoPNet to be human interpretable.

\citet{chen2018looks} suggest that we can interpret a classification decision by looking at the prototypes which the ProtoPNet found to be most strongly present in the input. For a particular prototype, this means that a user looks at the prototype's visualisation (i.e. a patch of a training image) and the patch in the input image which the ProtoPNet finds most similar to the given prototype. This is indeed similar to how humans communicate their decisions. However, it is only meaningful \emph{if} the other person finds two parts of an object similar for the same reasons. Otherwise, the decision of one person is not understandable to the other. Hence, without such a mutual understanding between a human and a ProtoPNet, displaying prototype similarities does not per-se yield interpretability. Consequently, Equivalence~\ref{eq:equivalence} needs to hold for a ProtoPNet to be interpretable.

Indeed, if a human and a ProtoPNet find two image patches similar for different reasons, interpretability would require the human to make sense of the latent representations which are the ProtoPNet's basis for scoring similarity. Yet, these representations are the outputs of a convolutional backbone and can hence not be considered interpretable.

\section{Implementation Details}
\label{app:implementation-details}
We implemented the ProtoPNets, as well as our adversarial attacks using PyTorch \cite{paszke2019pytorch}. We use exactly the same experimental setup as described in \citet{chen2018looks} for our experiments. Particularly, we use 40x augmentation for each training image in the CUB-200-2011 dataset\footnote{The CUB-200-2011 dataset can be downloaded from \url{ http://www.vision.caltech.edu/visipedia/CUB-200-2011.html}.}\cite{WahCUB_200_2011}. 

Following \citet{chen2018looks}, we set the number of channels in a prototype to $128$ for ResNet-18 and VGG-19; and $256$ for ResNet-34. Additionally, we set the prototype size to $1\times1$ and use $10$ prototypes for each class. Finally, we simply reuse their training setup for ease of implementation and reproducibility. We set the weights of cluster and separation costs to 0.8 and -0.08 respectively, and choose the coefficient of the $L_1$ regularization as $0.0001$. For training, we set the number of warmup epochs to $5$ and use Adam optimizer with an initial learning rate of $0.003$. Subsequently, we train the entire model for 6 more epochs with a learning rate of $0.0001$ for the backbone network and $0.003$ for the prototype layer. Lastly, we push the prototypes to the most similar latent training image patches and fine-tune the fully-connected layer for $20$ iterations using Adam optimizer with a learning rate of $0.0001$. We select the best performing model on the test set for our experiments. The code to our experiments is openly available at \url{https://github.com/fanconic/this-does-not-look-like-that}.


\begin{figure*} [bh]
\begin{subfigure}{0.32\textwidth}
  \centering
  \includegraphics[width=\linewidth]{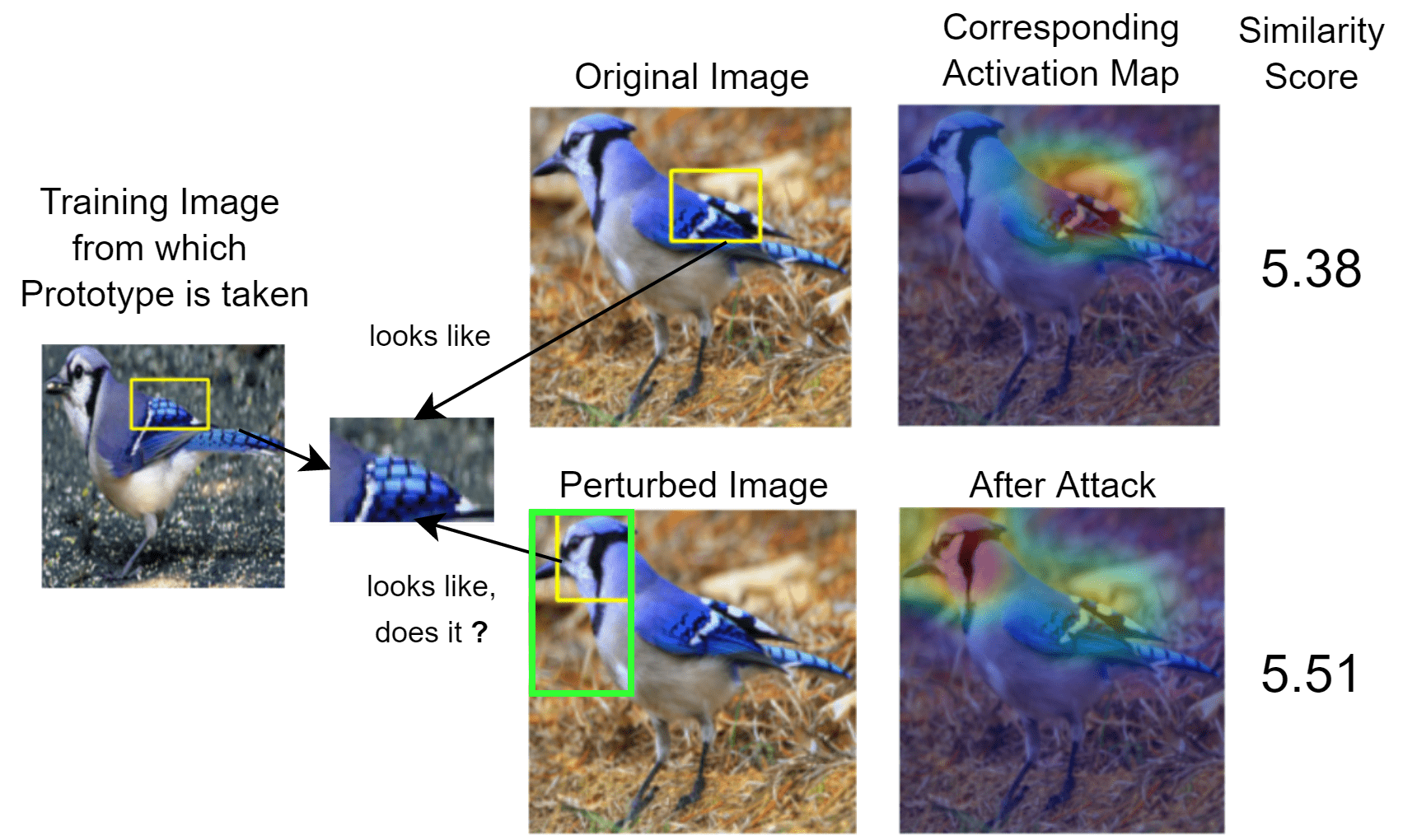}\\
  \caption{}
\end{subfigure}
\begin{subfigure}{0.32\textwidth}
  \centering
  \includegraphics[width=\linewidth]{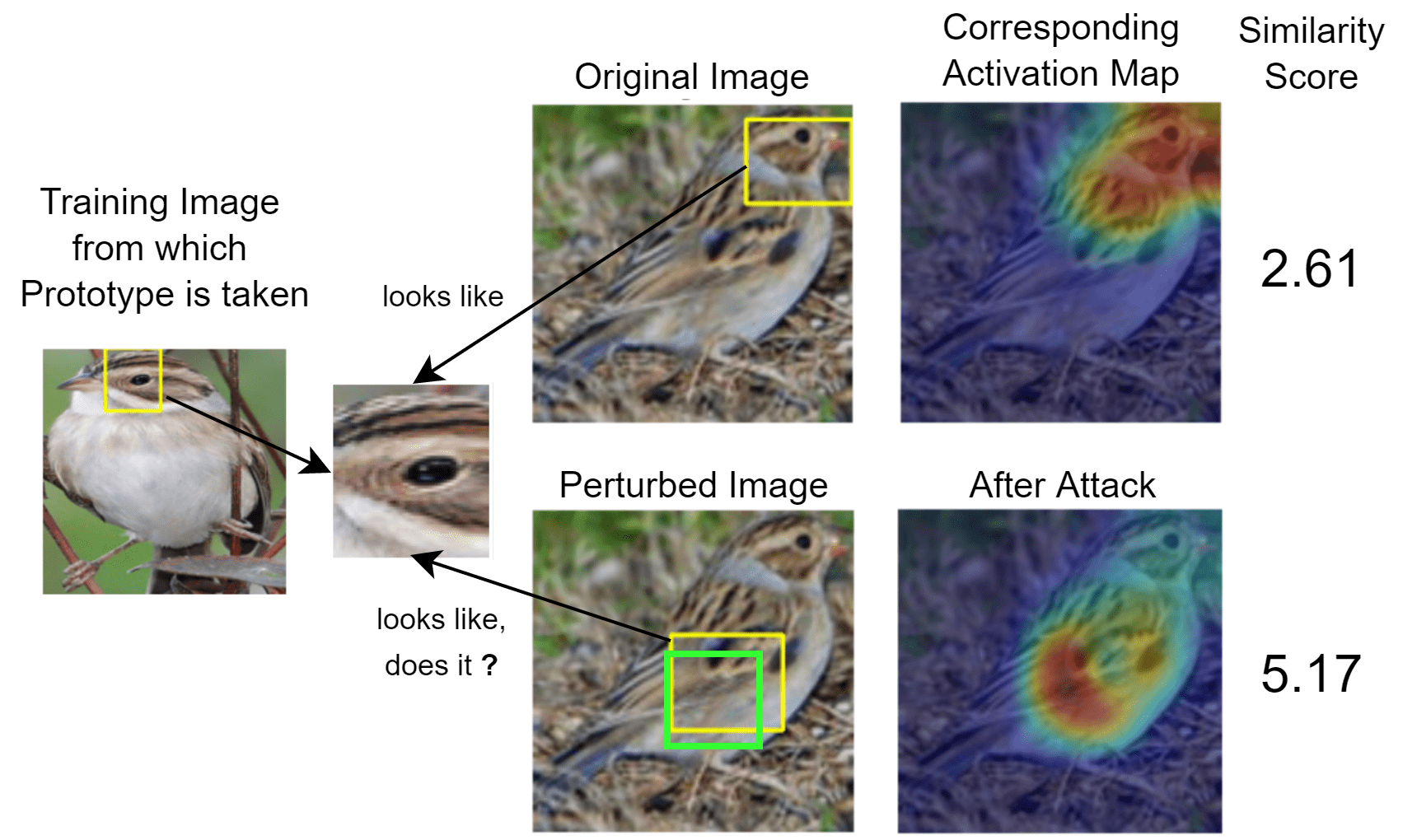}
  \caption{}
\end{subfigure}\hfill
\begin{subfigure}{0.32\textwidth}
  \centering
  \includegraphics[width=\linewidth]{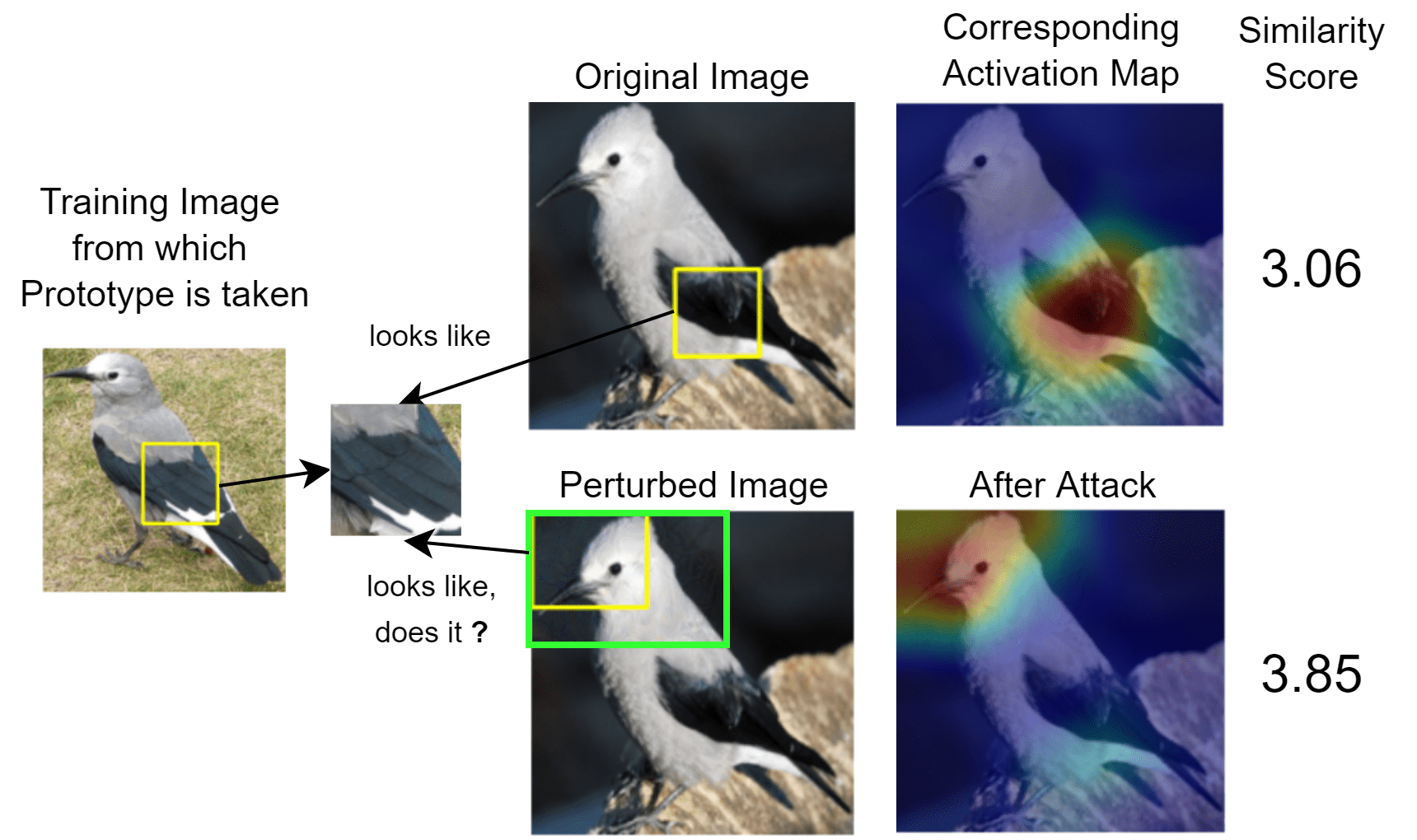}
  \caption{}
\end{subfigure}\hfill
\begin{subfigure}{0.32\textwidth}
  \centering
  \includegraphics[width=\linewidth]{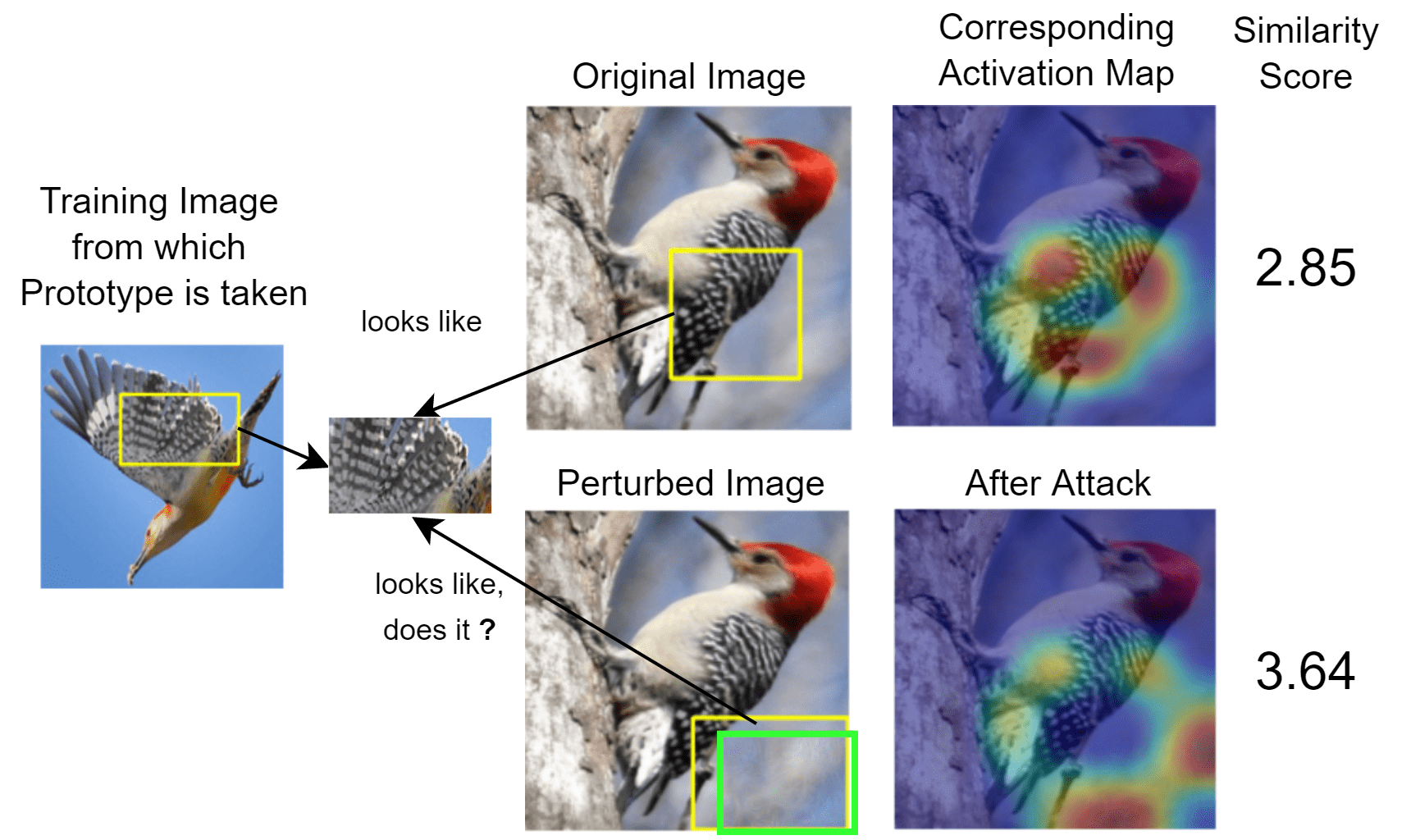}
  \caption{}
\end{subfigure}\hfill
\begin{subfigure}{0.32\textwidth}
  \centering
  \includegraphics[width=\linewidth]{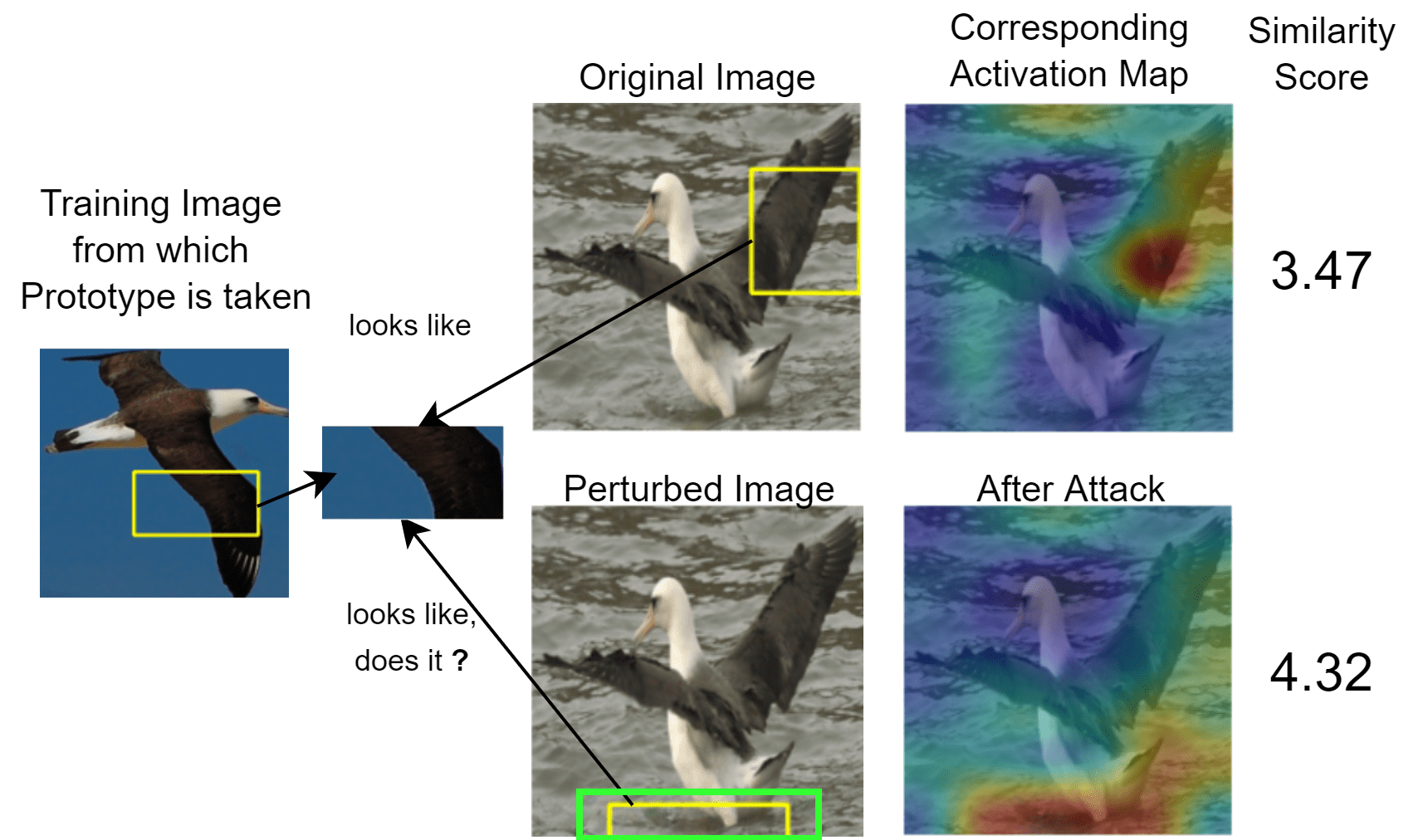}
  \caption{}
\end{subfigure}\hfill
\begin{subfigure}{0.32\textwidth}
  \centering
  \includegraphics[width=\linewidth]{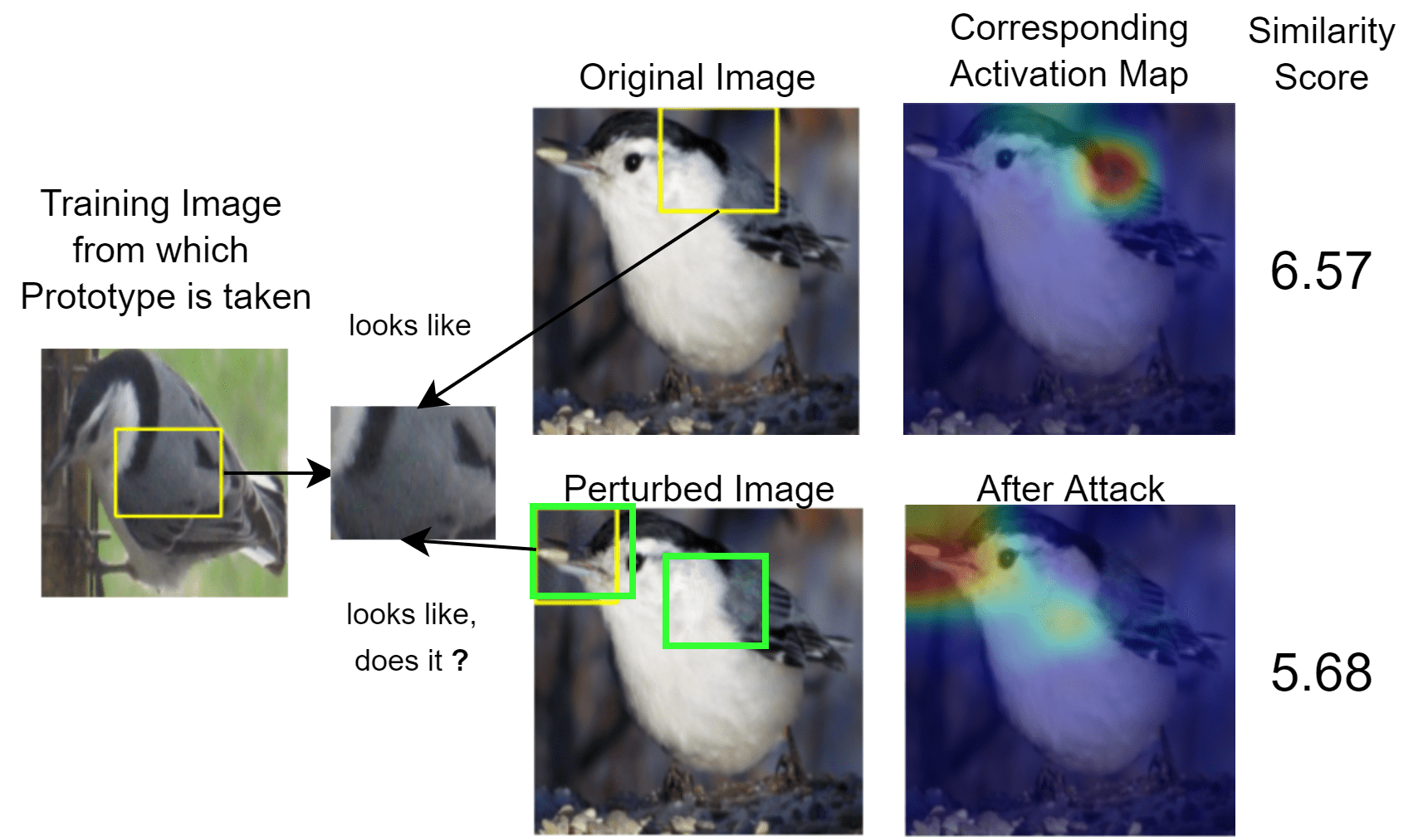}
  \caption{}
\end{subfigure}\hfill
\vspace{-1mm}
\caption{The Head on Stomach experiment - ResNet-18}
\label{fig:results_adv_rn18}
\end{figure*}

\section{More Examples for Head on Stomach Experiment} \label{app:hosexamples}

\begin{figure*}[h]
\begin{subfigure}{0.32\textwidth}
  \centering
  \includegraphics[width=\linewidth]{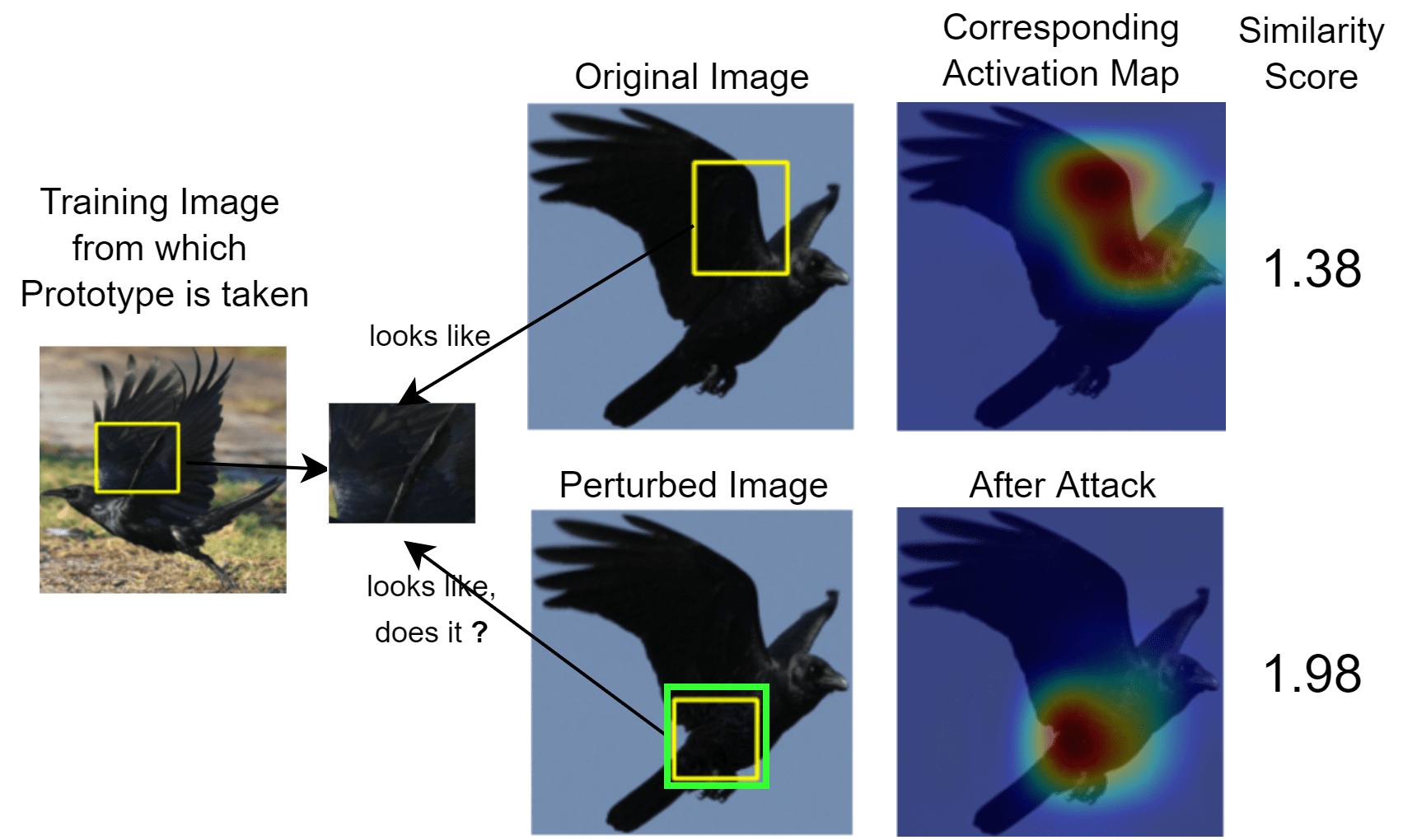}
  \caption{}
\end{subfigure}\hfill
\begin{subfigure}{0.32\textwidth}
  \centering
  \includegraphics[width=\linewidth]{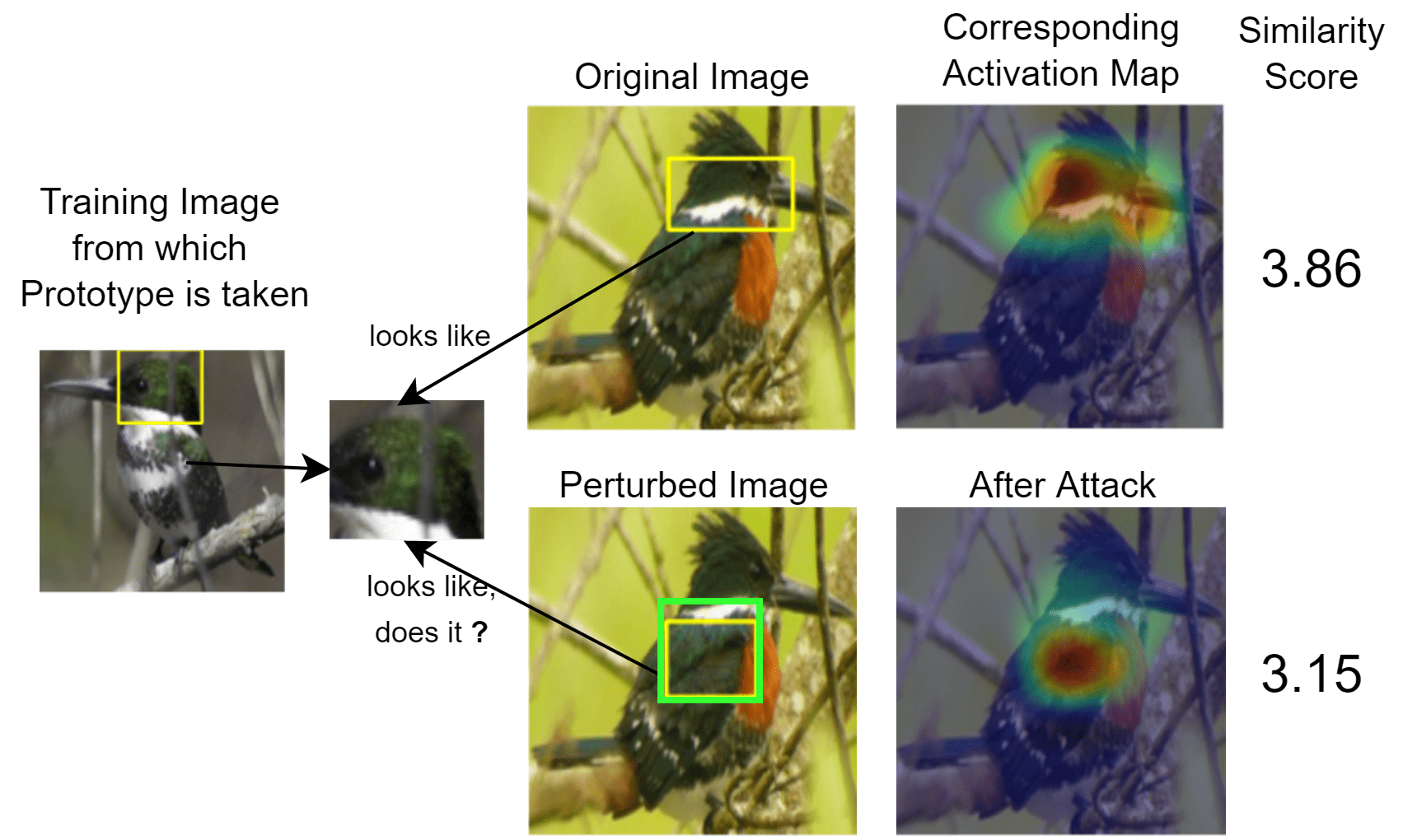}
  \caption{}
\end{subfigure}\hfill
\begin{subfigure}{0.32\textwidth}
  \centering
  \includegraphics[width=\linewidth]{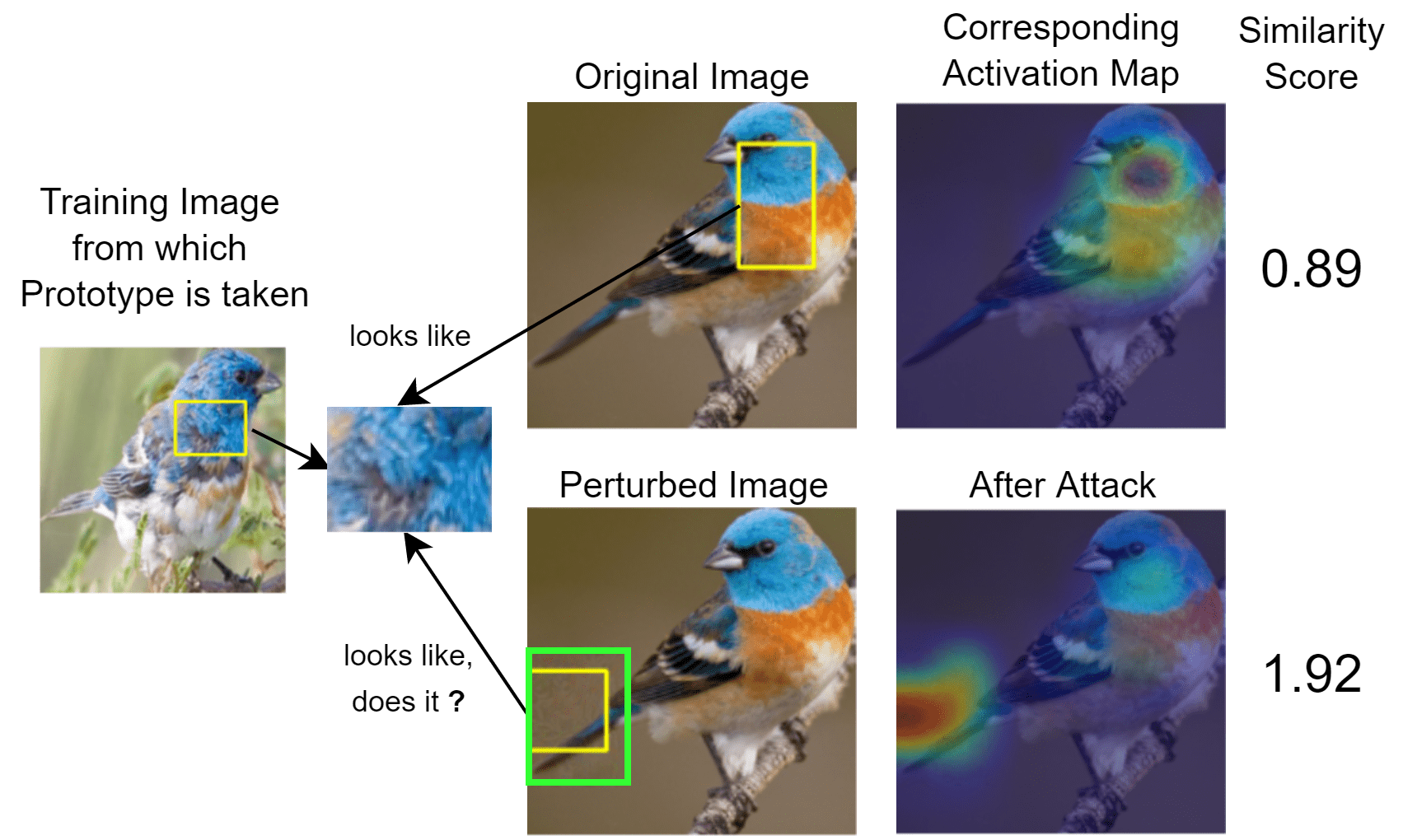}
  \caption{}
\end{subfigure}\hfill
\begin{subfigure}{0.32\textwidth}
  \centering
  \includegraphics[width=\linewidth]{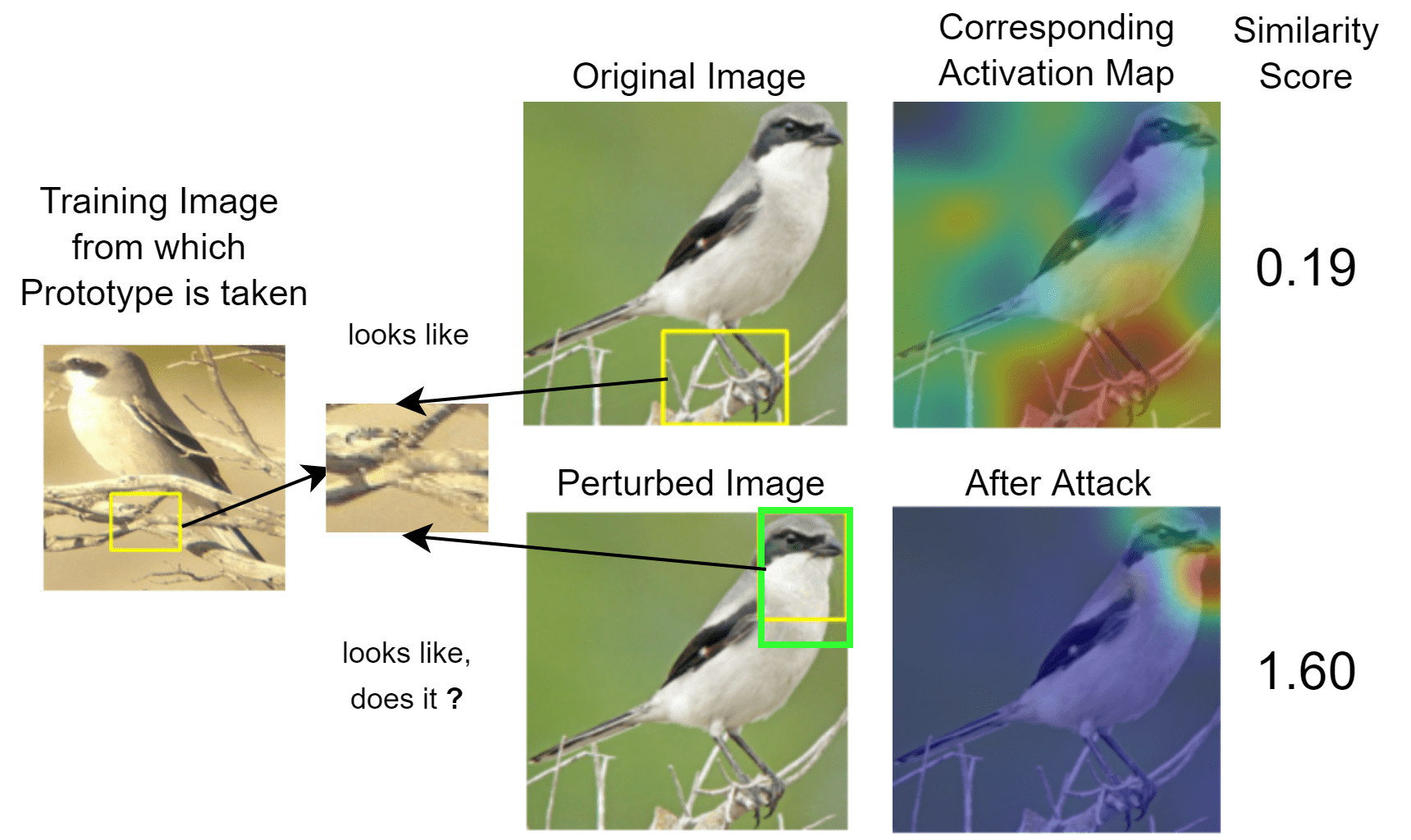}
  \caption{}
\end{subfigure}\hfill
\begin{subfigure}{0.32\textwidth}
  \centering
  \includegraphics[width=\linewidth]{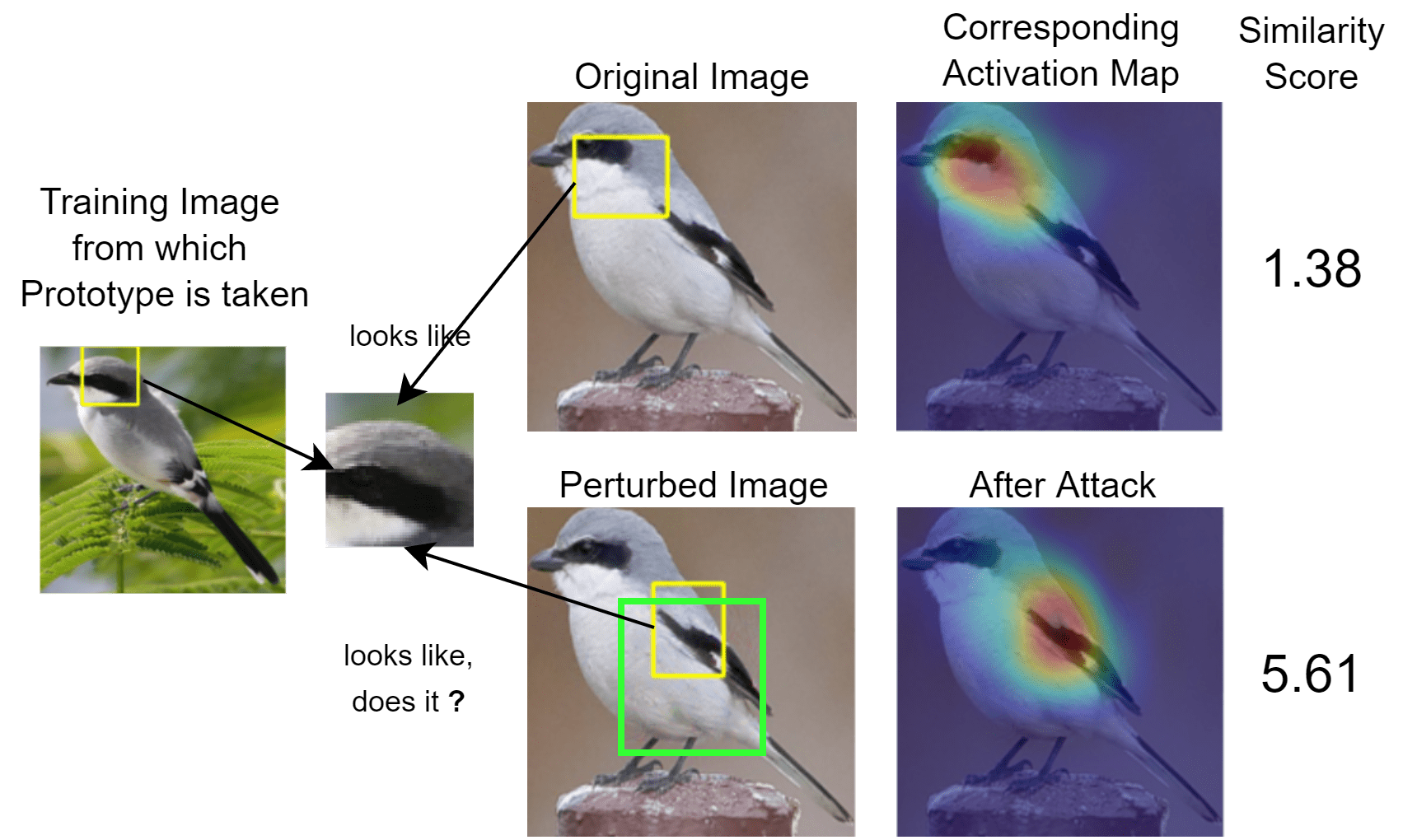}
  \caption{}
\end{subfigure}\hfill
\begin{subfigure}{0.32\textwidth}
  \centering
  \includegraphics[width=\linewidth]{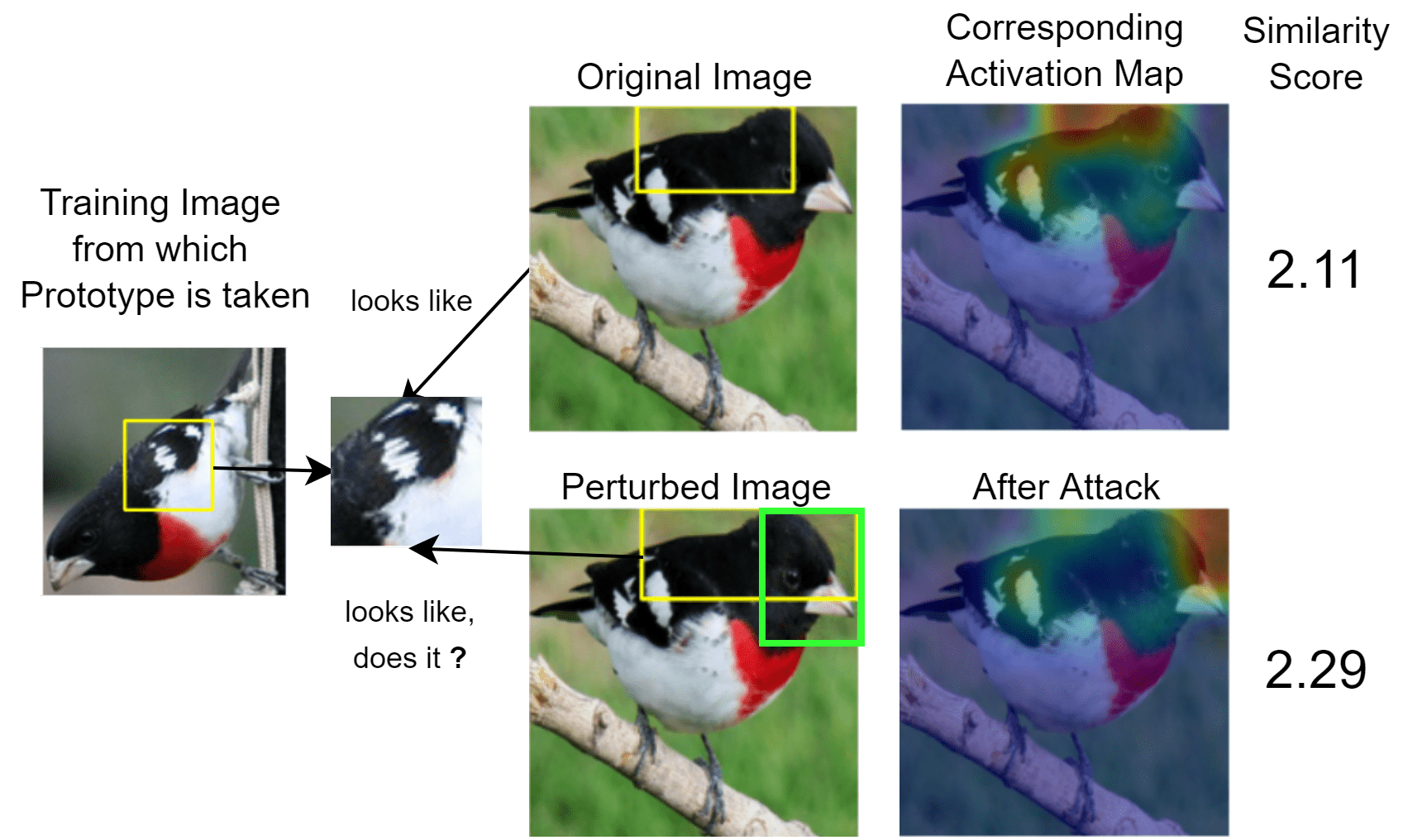}  \caption{}
\end{subfigure}\hfill
\vspace{-1mm}
\caption{The Head on Stomach experiment - ResNet-34}
\label{fig:results_adv_rn34}
\end{figure*}

\begin{figure*}[h]
\begin{subfigure}{0.32\textwidth}
  \centering
  \includegraphics[width=\linewidth]{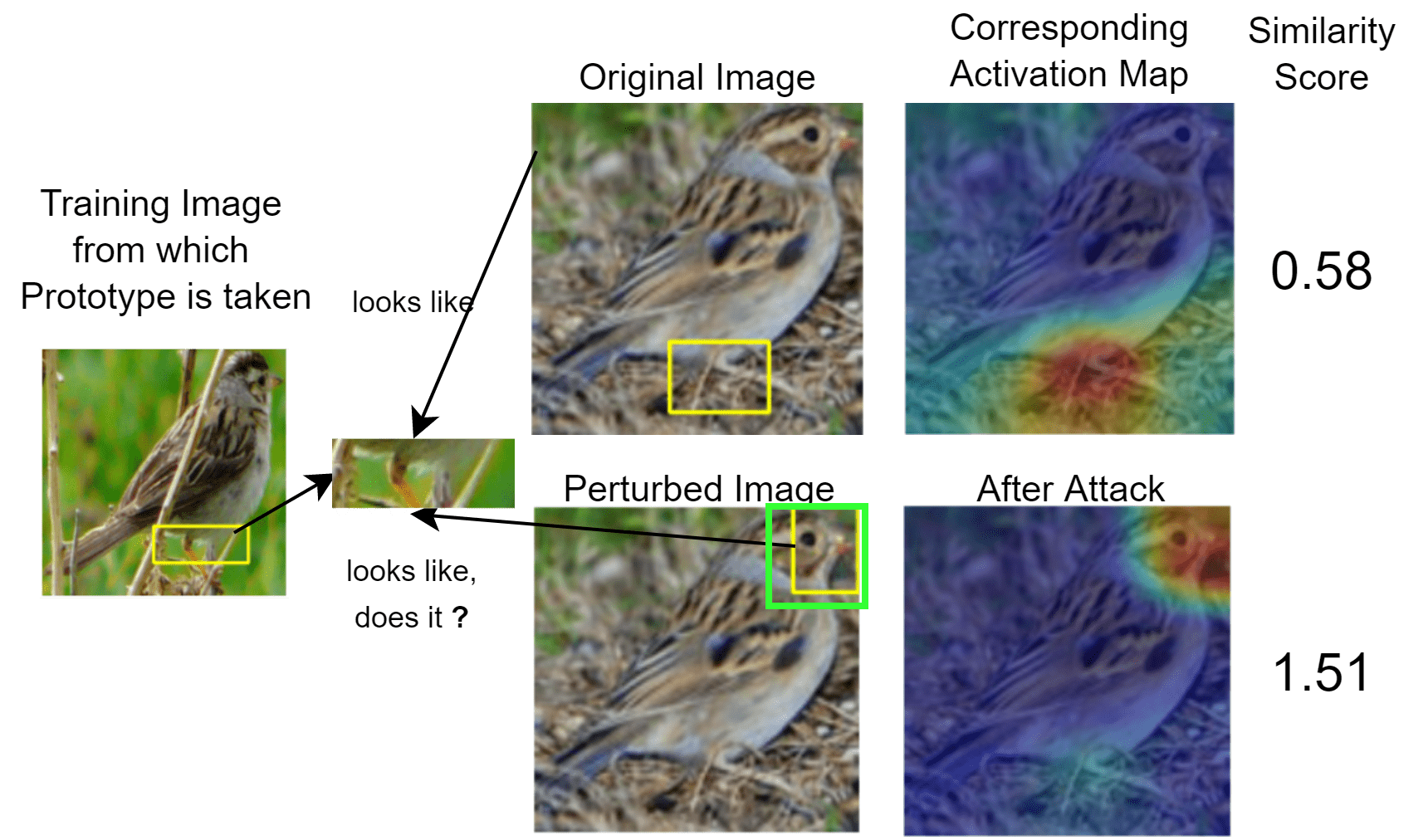}  \caption{}
\end{subfigure}\hfill
\begin{subfigure}{0.32\textwidth}
  \centering
  \includegraphics[width=\linewidth]{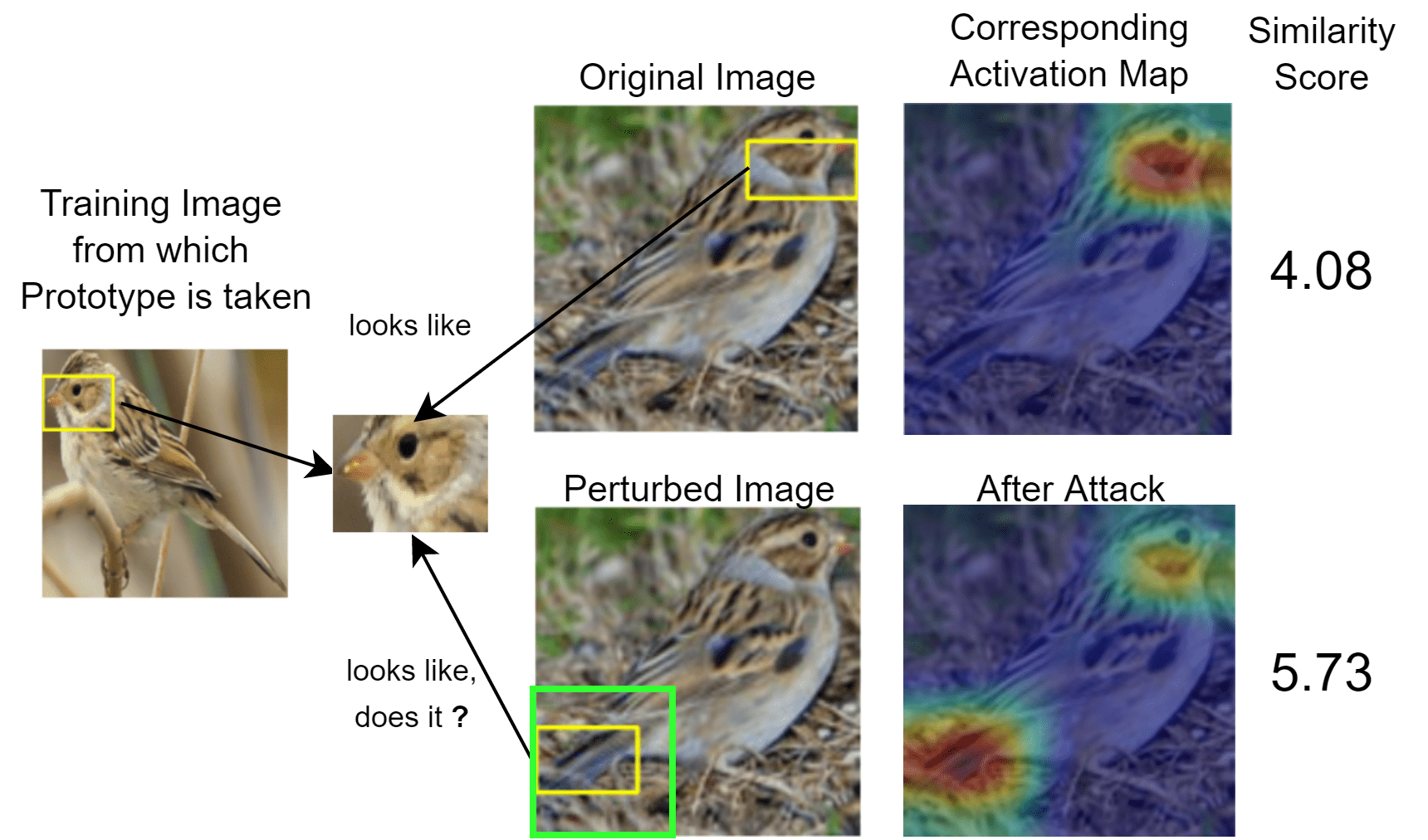}  \caption{}
\end{subfigure}\hfill
\begin{subfigure}{0.32\textwidth}
  \centering
  \includegraphics[width=\linewidth]{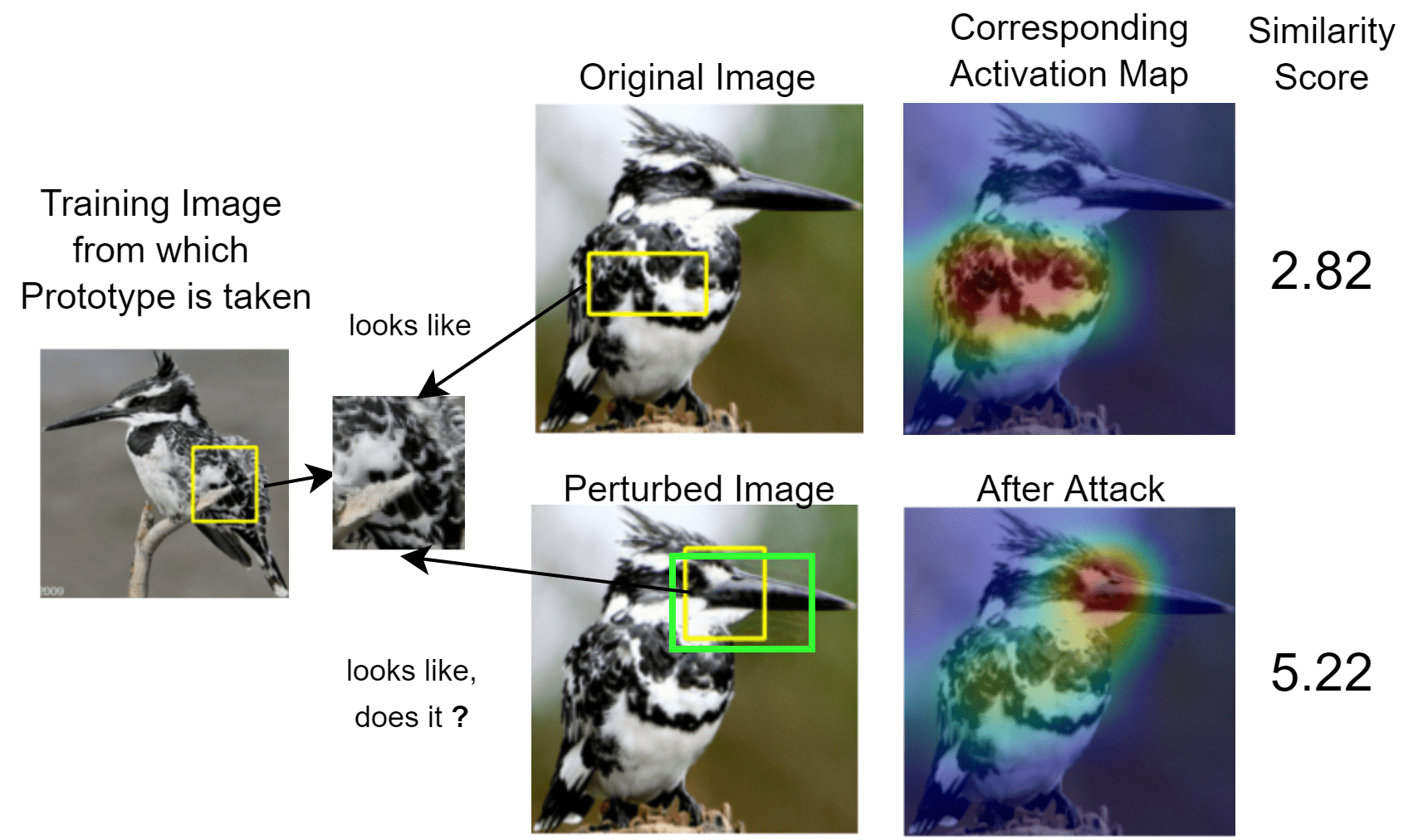}  \caption{}
\end{subfigure}\hfill
\begin{subfigure}{0.32\textwidth}
  \centering
  \includegraphics[width=\linewidth]{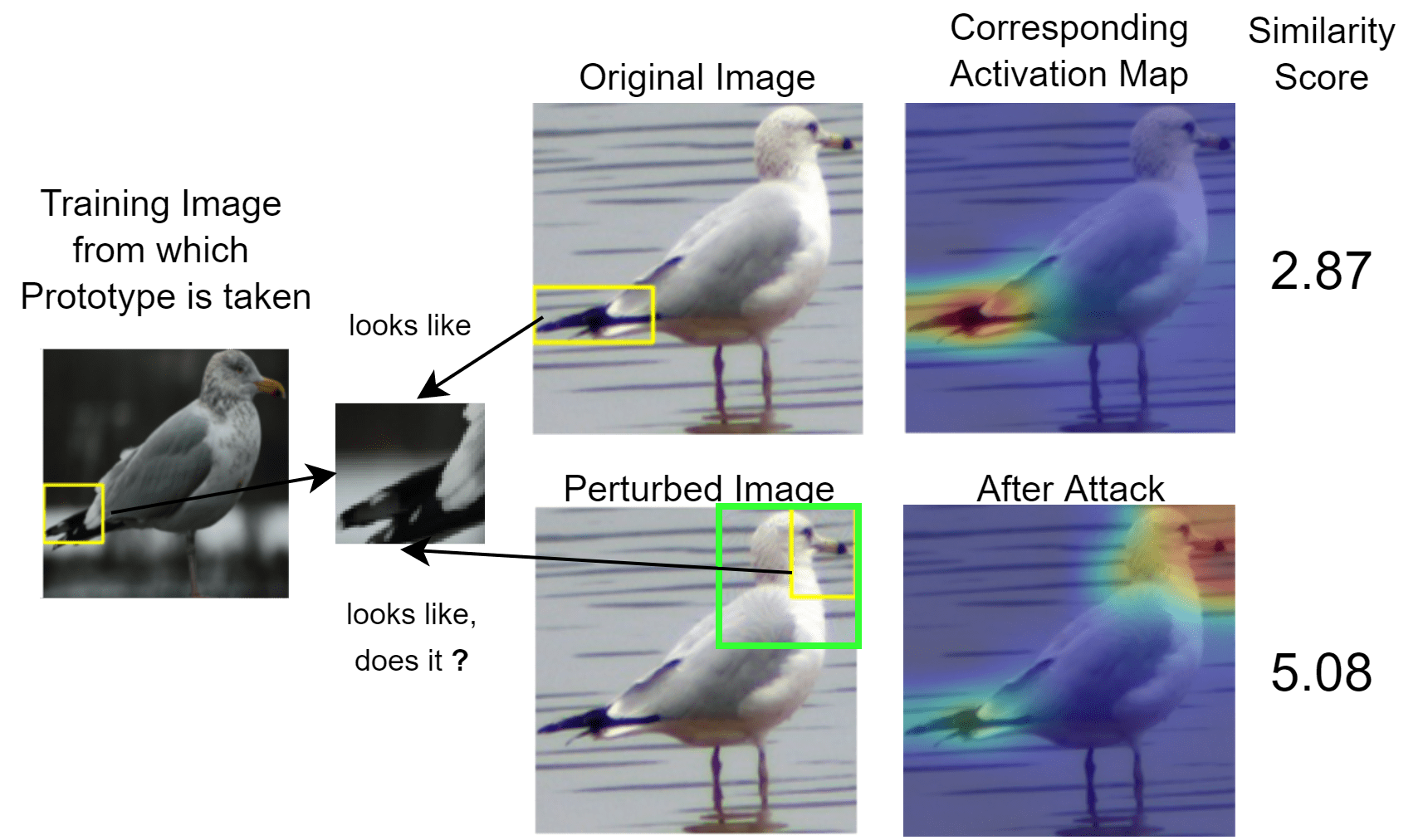}  \caption{}
\end{subfigure}\hfill
\begin{subfigure}{0.32\textwidth}
  \centering
  \includegraphics[width=\linewidth]{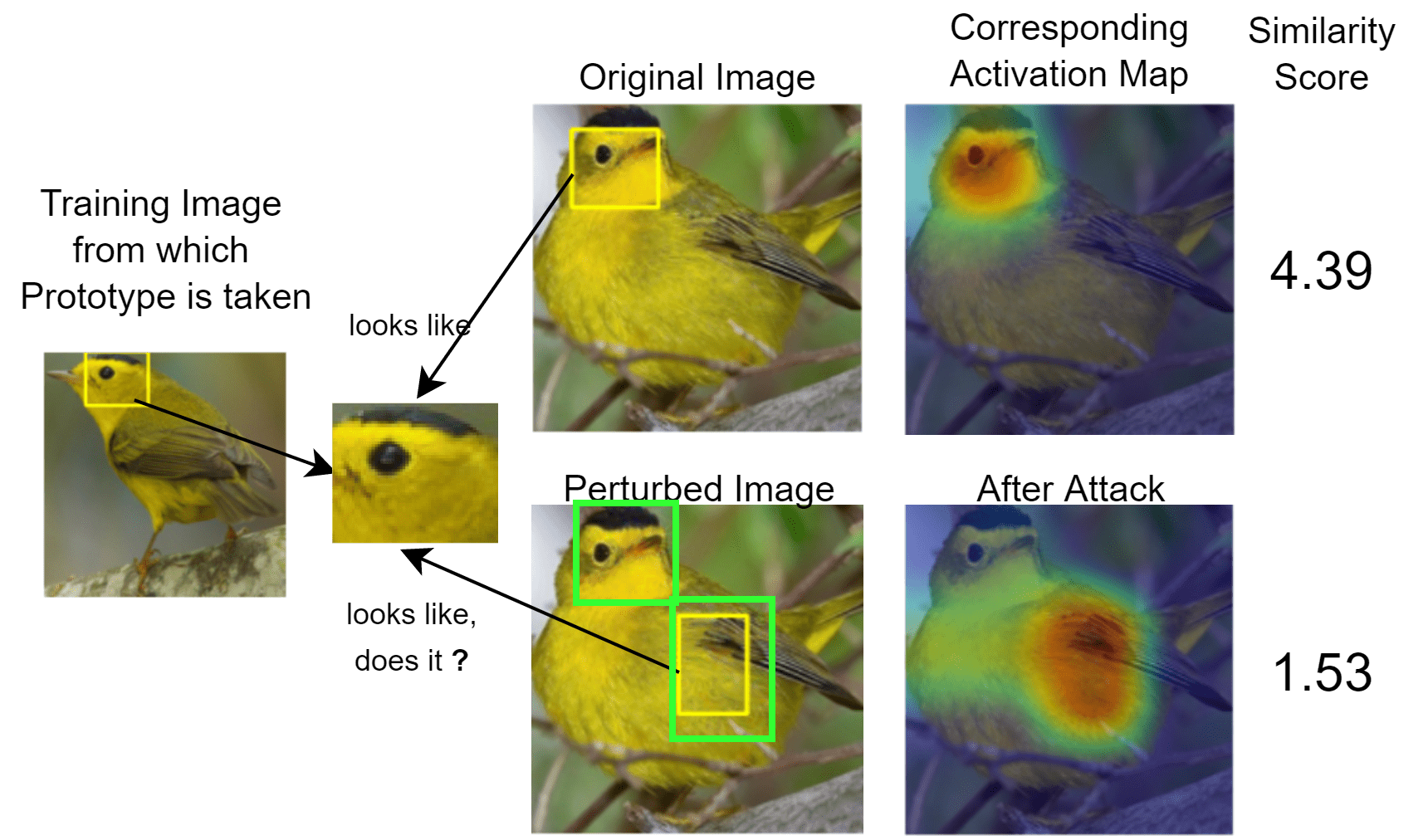}  \caption{}
\end{subfigure}\hfill
\begin{subfigure}{0.32\textwidth}
  \centering
  \includegraphics[width=\linewidth]{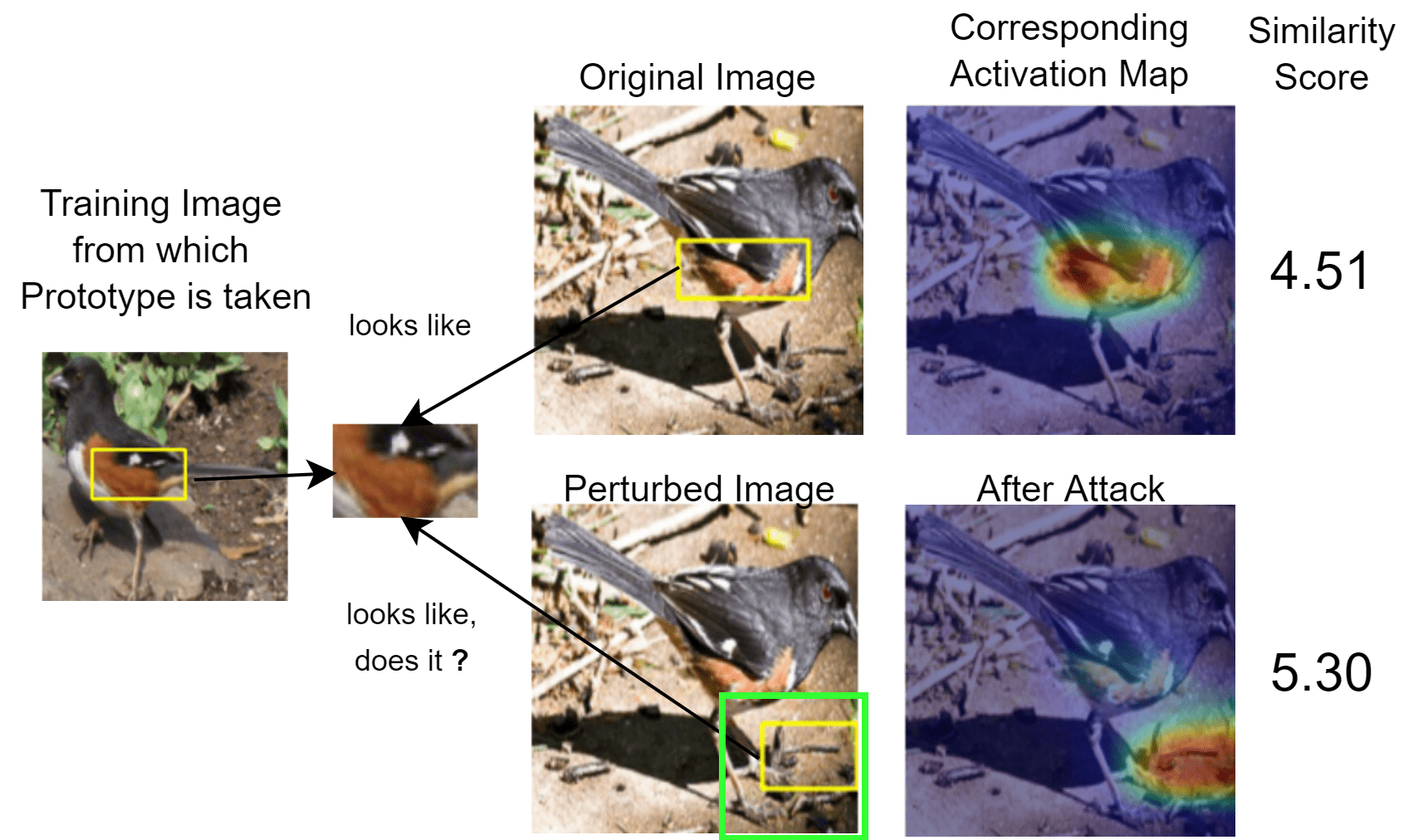}  \caption{}
\end{subfigure}\hfill
\vspace{-1mm}
\caption{The Head on Stomach experiment - VGG-19}
\label{fig:results_adv_vgg19}
\end{figure*}

In this section, we present further examples from the Head on Stomach experiment which were successful. Figures~\ref{fig:results_adv_rn18},~\ref{fig:results_adv_rn34}~and~\ref{fig:results_adv_vgg19} illustrate the results for ResNet-18, ResNet-34 and VGG-19 respectively. As evident, after perturbing the images with carefully crafted noise, the interpretations provided by ProtoPNets become non-sensical for humans, thereby indicating that the ProtoPNet's notion of similarity does not align with that of humans.

\FloatBarrier
\newpage
\section{Details on JPEG Experiment}\label{app:jpeg}

\begin{figure}[h]
  \centering
    \begin{minipage}[c]{0.65\textwidth}
      \centering
      \minipage{0.48\textwidth}
        \centering
        \includegraphics[width=0.95\linewidth]{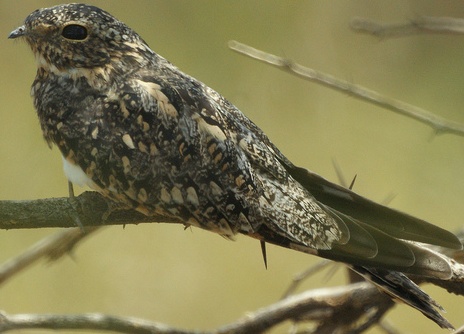} \\ original
      \endminipage\hfill
      \minipage{0.48\textwidth}
        \centering
        \includegraphics[width=0.95\linewidth]{figures/compression  visualisation/Nighthawk_0066_82238_compressed_20.JPEG}\\compressed
      \endminipage\hfill
    \end{minipage}
    \caption{A random image from the the CUB-200-2011 dataset without and with compression artefacts respectively (JPEG compression quality is set to 20\%) }\label{fig:jpeg_compression}
\end{figure}

\subsection{More Examples for JPEG Experiment} \label{app:jpegexamples}
\begin{figure*} [h]
\begin{subfigure}{0.32\textwidth}
  \centering
  \includegraphics[width=\linewidth]{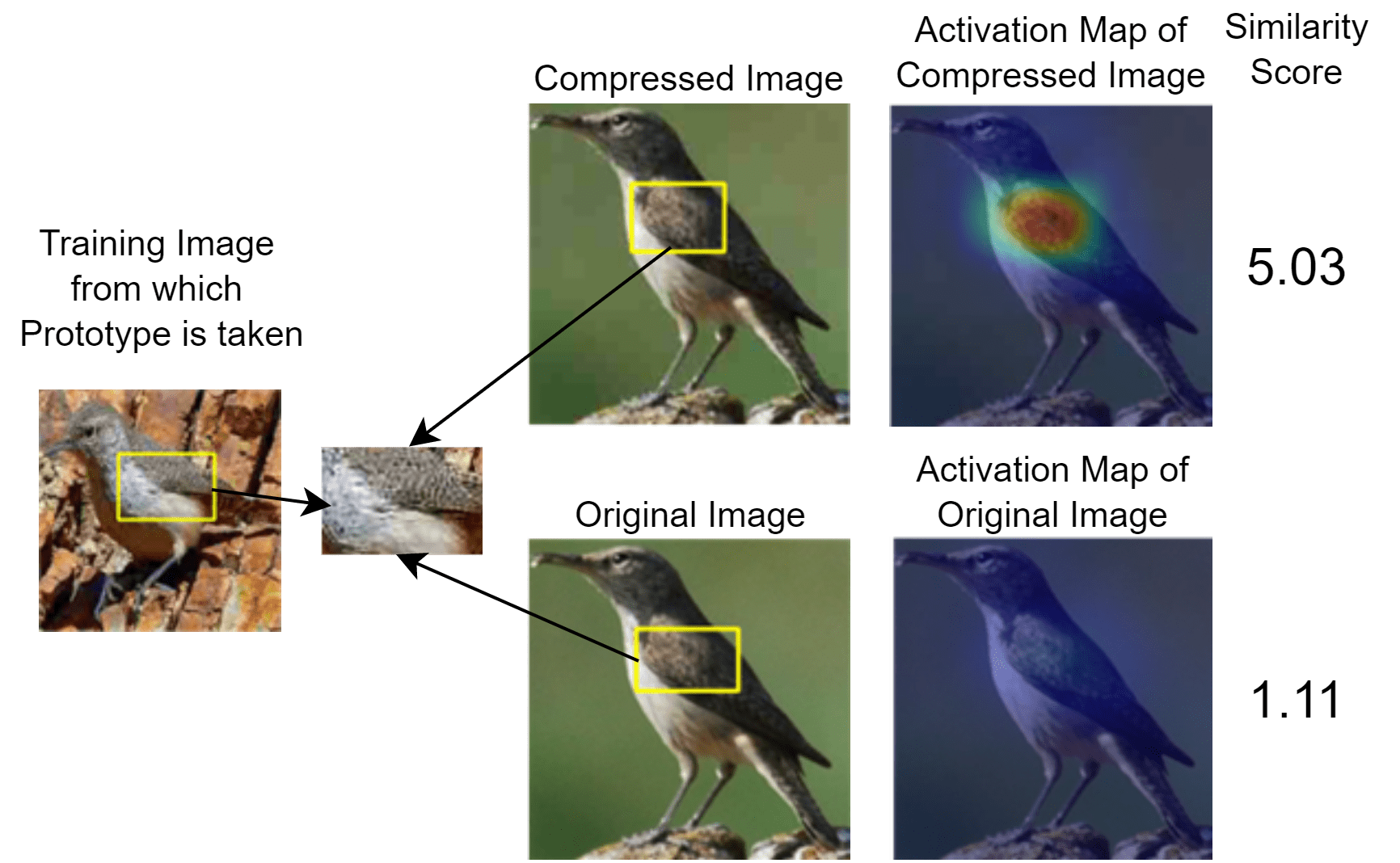}  
  \caption{}
\end{subfigure}\hfill
\begin{subfigure}{0.32\textwidth}
  \centering
  \includegraphics[width=\linewidth]{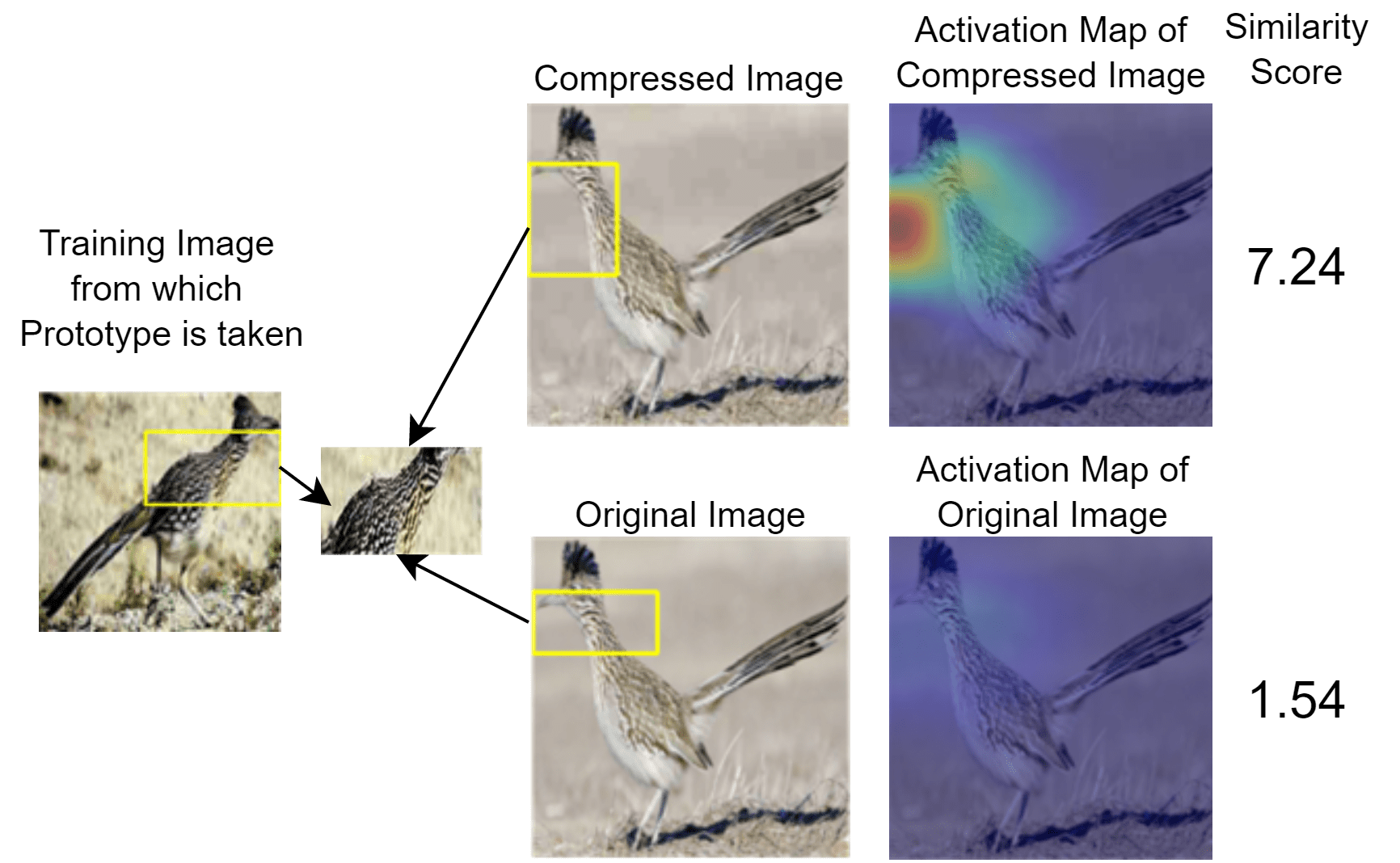}  
  \caption{}
\end{subfigure}\hfill
\begin{subfigure}{0.32\textwidth}
\centering
  \includegraphics[width=\linewidth]{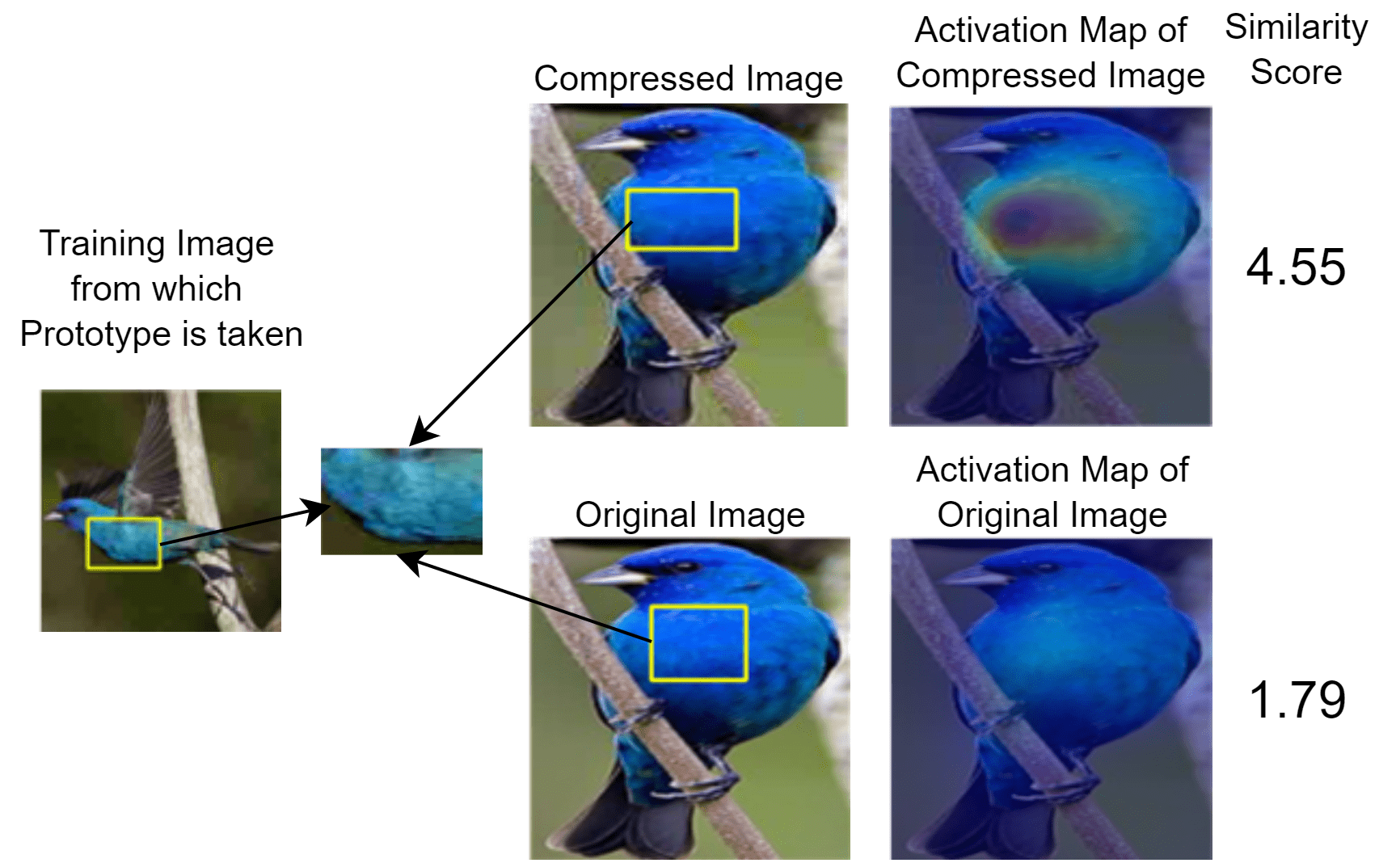}
  \caption{}
\end{subfigure}\hfill
\begin{subfigure}{0.32\textwidth}
  \centering
  \includegraphics[width=\linewidth]{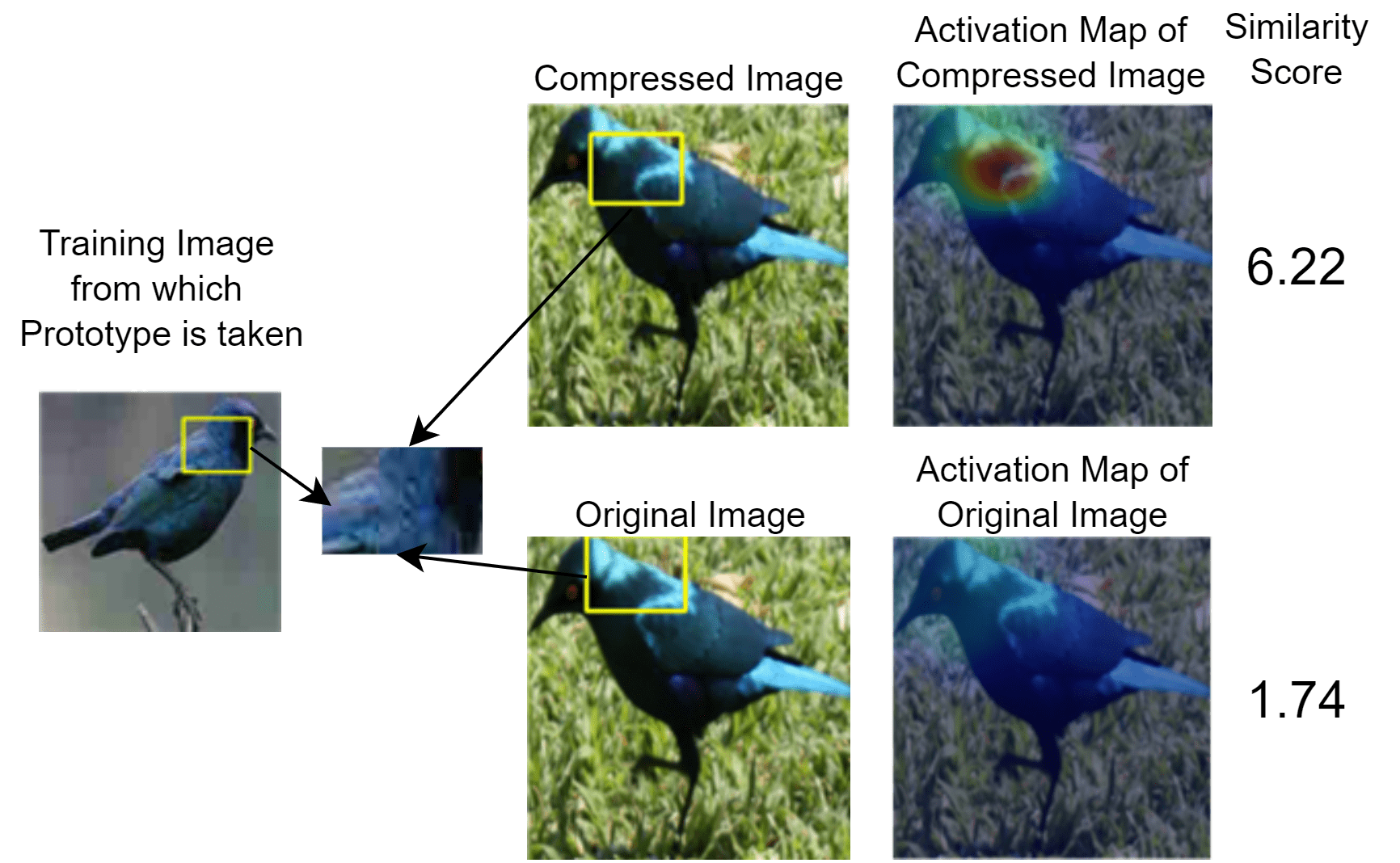}
  \caption{}
\end{subfigure}\hfill
\begin{subfigure}{0.32\textwidth}
  \centering
  \includegraphics[width=\linewidth]{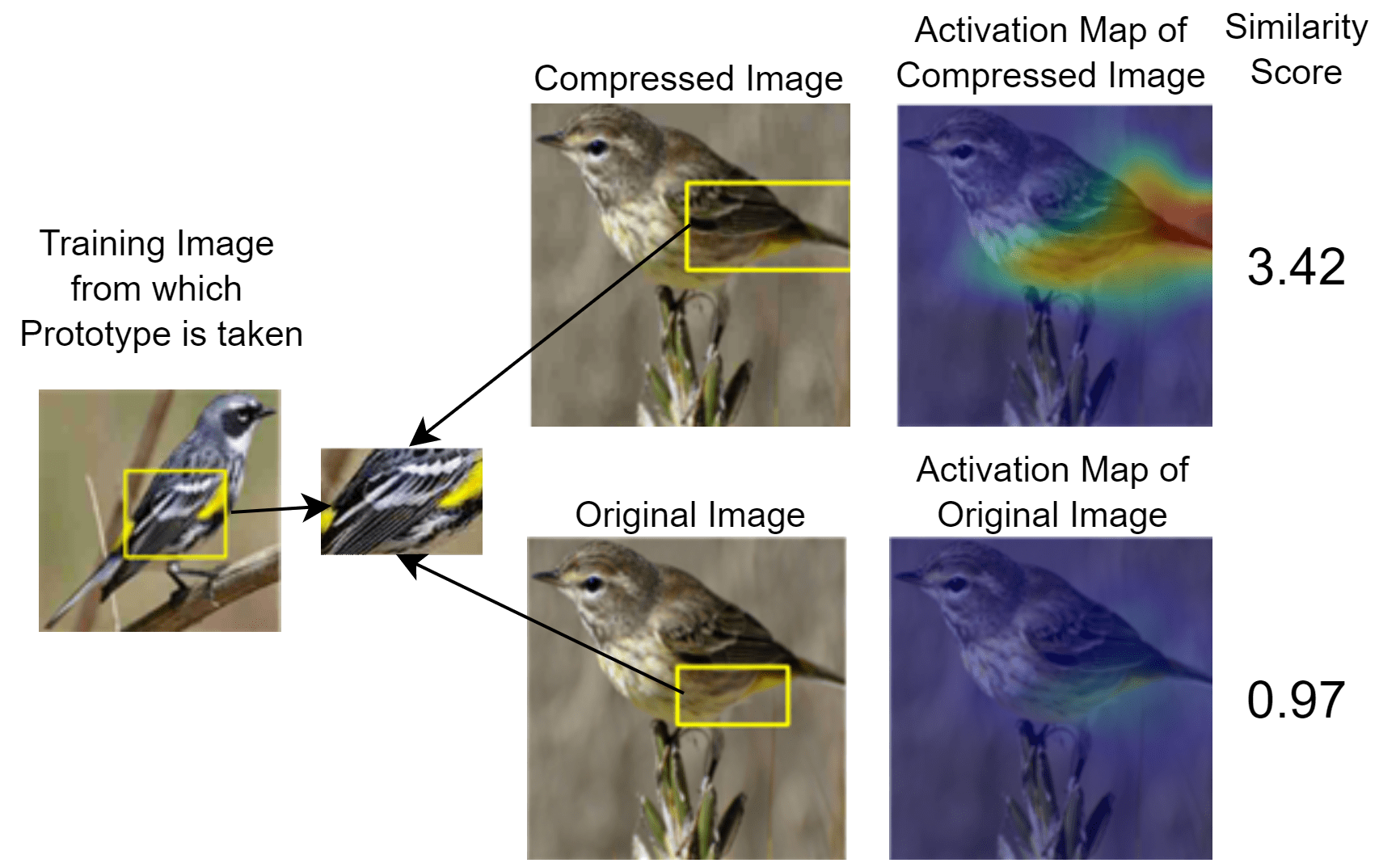}
  \caption{}
\end{subfigure}\hfill
\begin{subfigure}{0.32\textwidth}
  \centering
  \includegraphics[width=\linewidth]{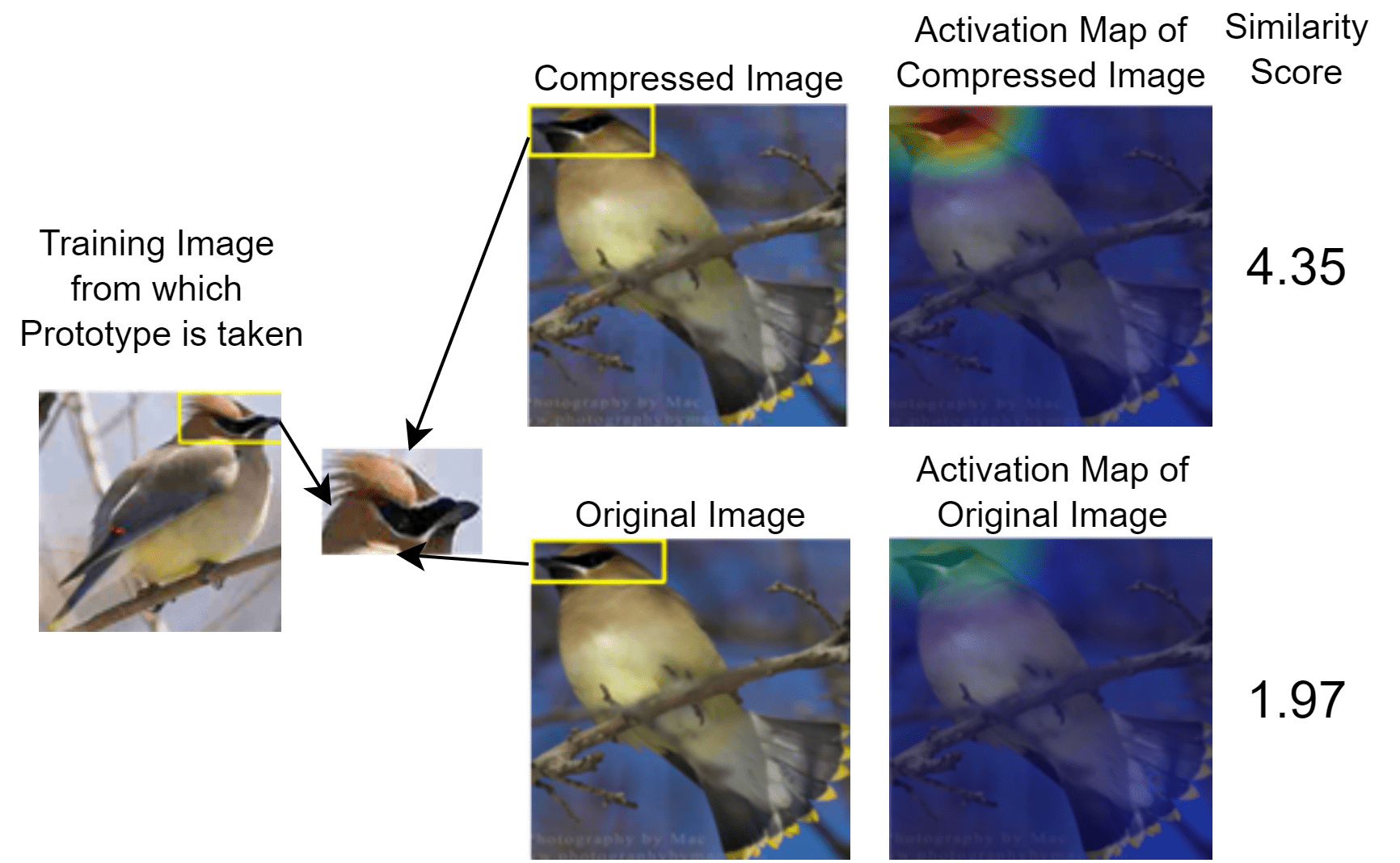}
  \caption{}
\end{subfigure}\hfill
\vspace{-1mm}
\caption{The JPEG experiment - ResNet-18}
\label{fig:results_jpeg_rn18}
\end{figure*}

\begin{figure*} [h]
\begin{subfigure}{0.32\textwidth}
  \centering
  \includegraphics[width=\linewidth]{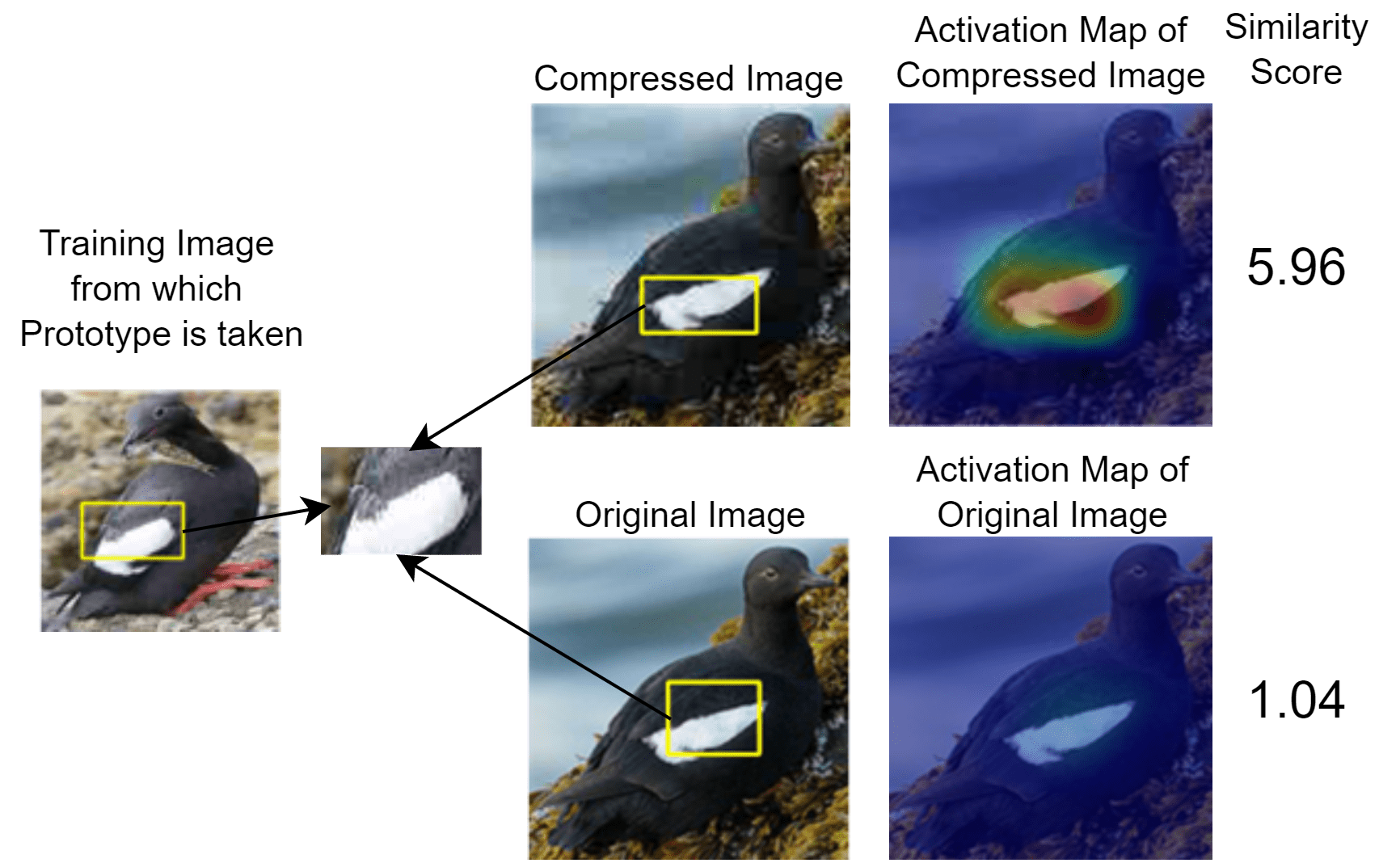}
  \caption{}
\end{subfigure}\hfill
\begin{subfigure}{0.32\textwidth}
  \centering
  \includegraphics[width=\linewidth]{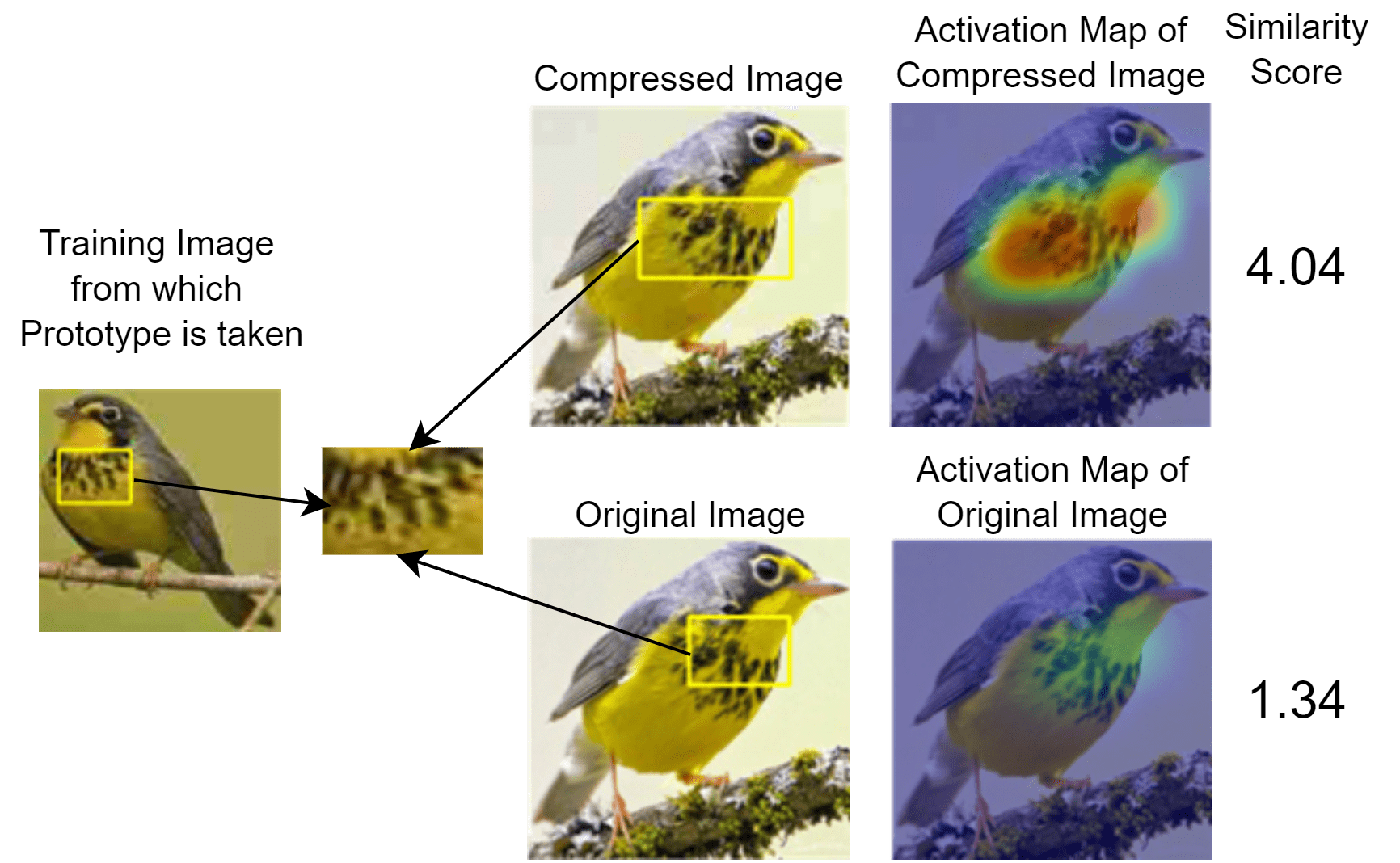}
  \caption{}
\end{subfigure}\hfill
\begin{subfigure}{0.32\textwidth}
  \centering
  \includegraphics[width=\linewidth]{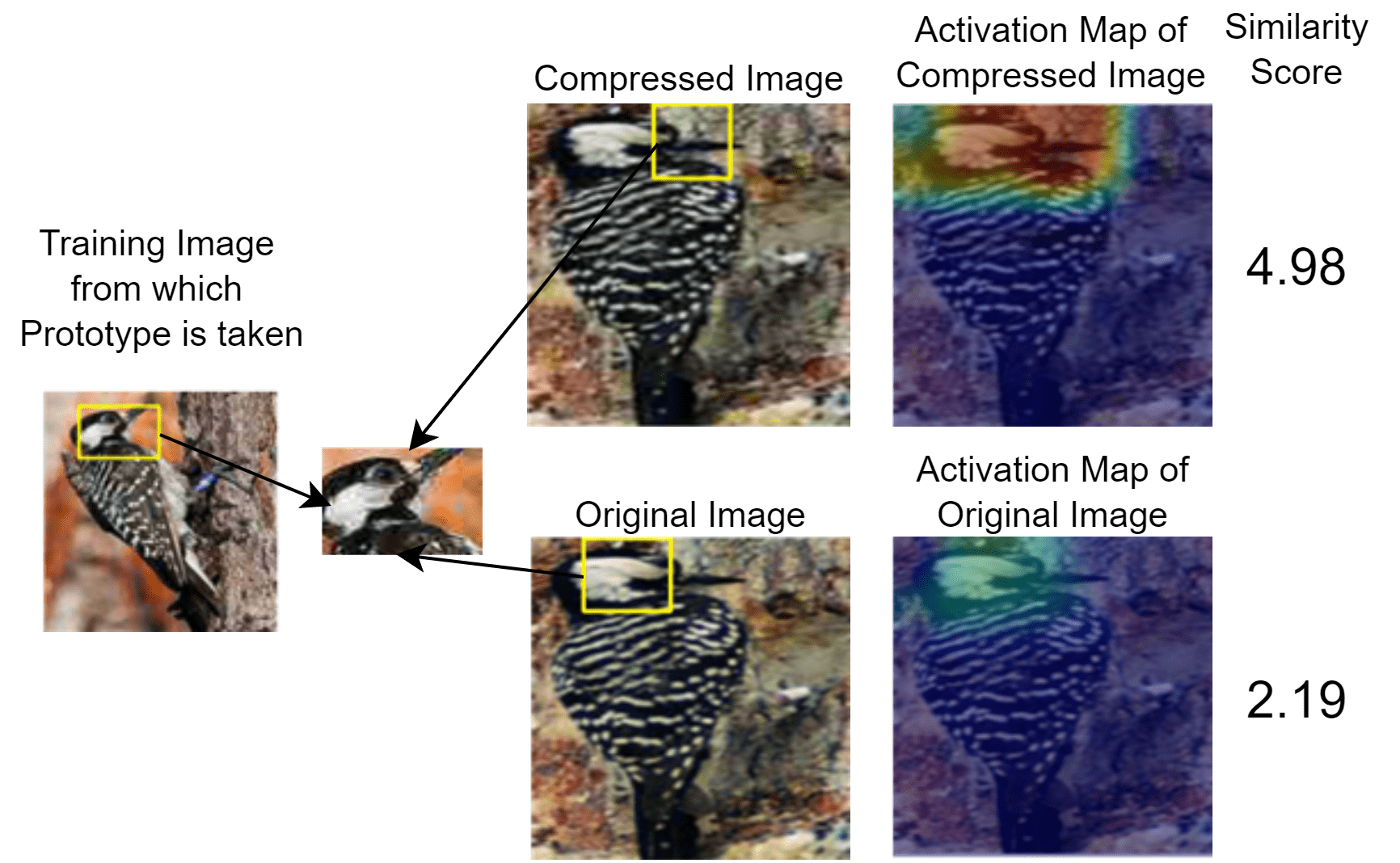}
  \caption{}
\end{subfigure}\hfill
\begin{subfigure}{0.32\textwidth}
  \centering
  \includegraphics[width=\linewidth]{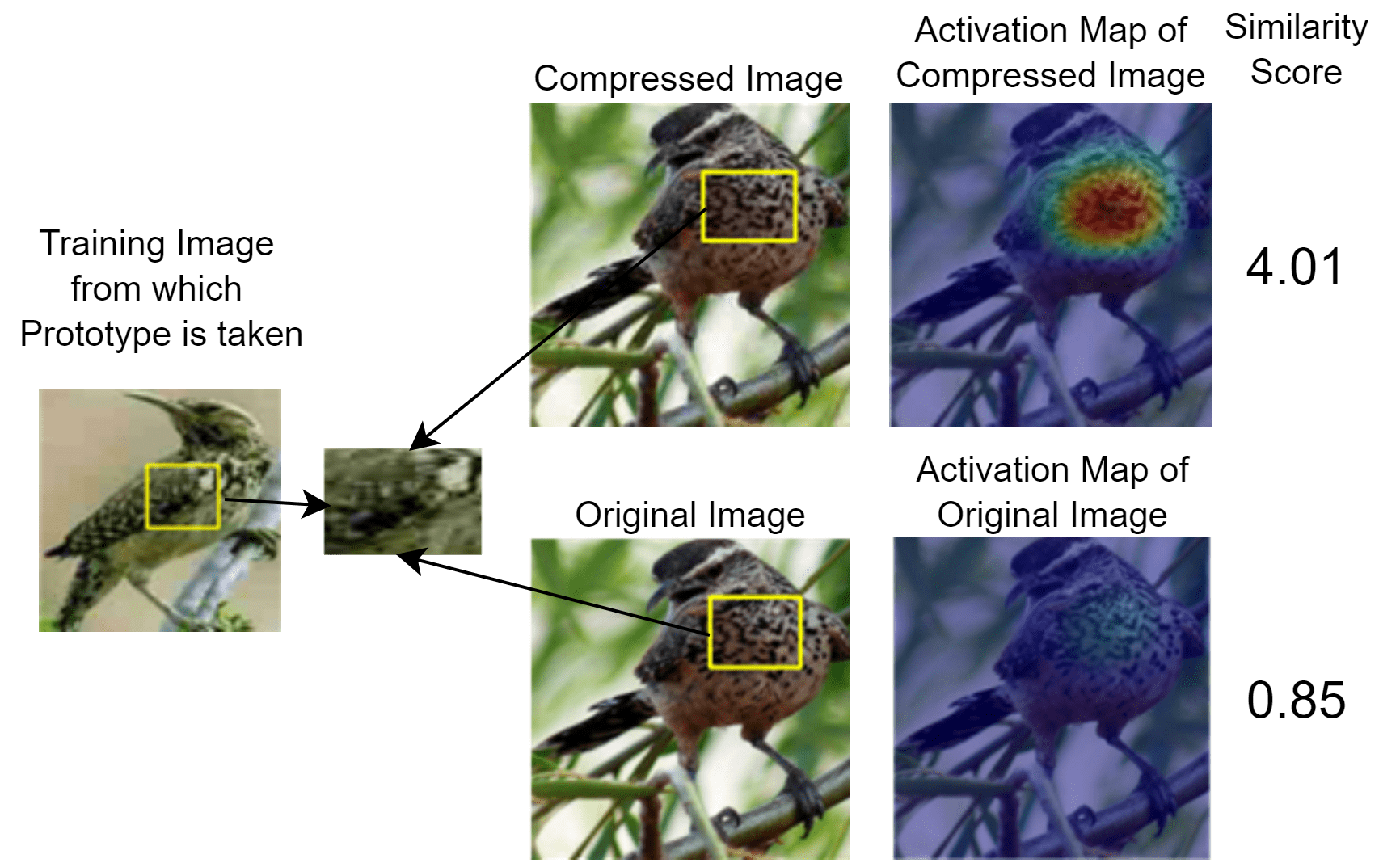}
  \caption{}
\end{subfigure}\hfill
\begin{subfigure}{0.32\textwidth}
  \centering
  \includegraphics[width=\linewidth]{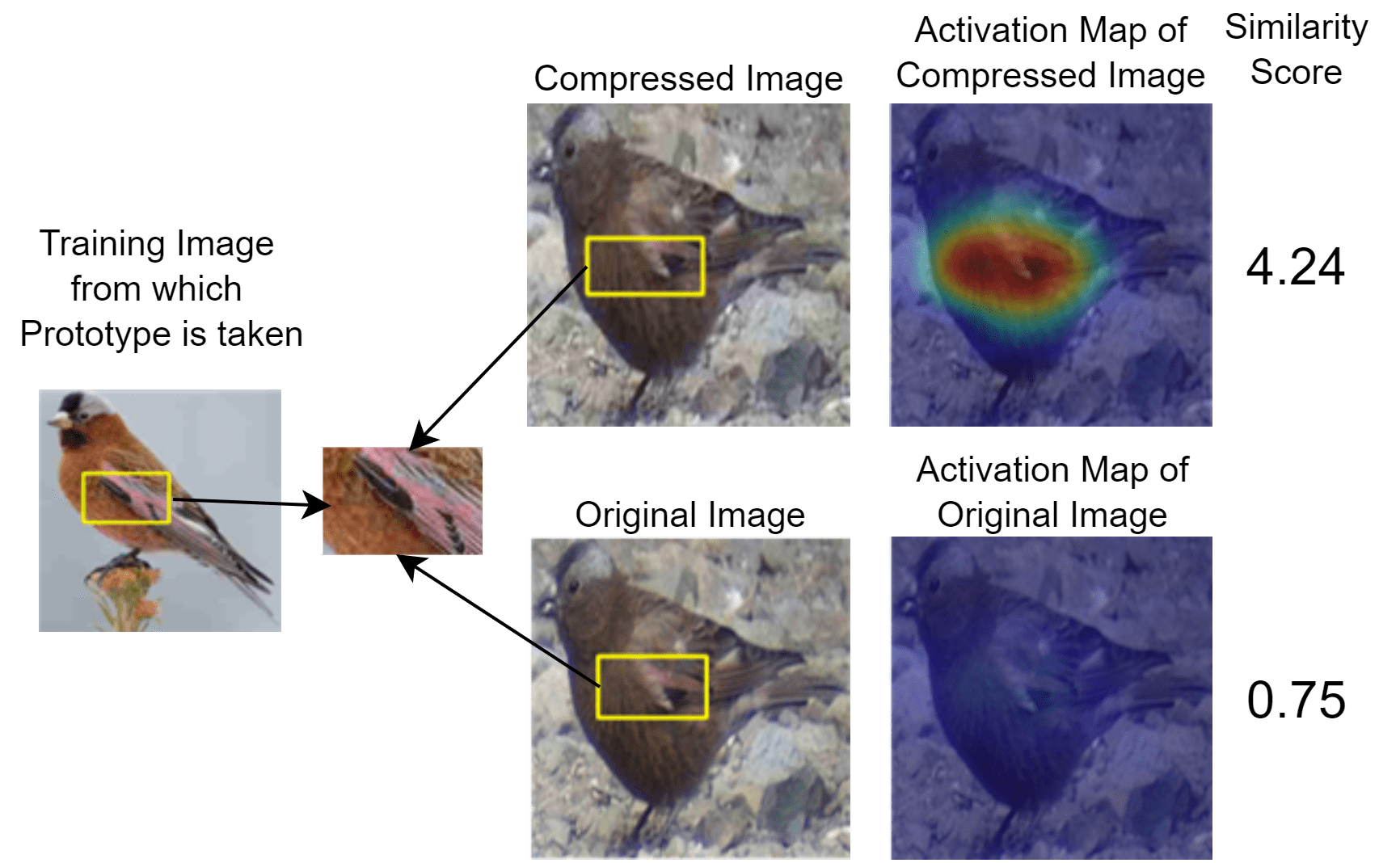}
  \caption{}
\end{subfigure}\hfill
\begin{subfigure}{0.32\textwidth}
  \centering
  \includegraphics[width=\linewidth]{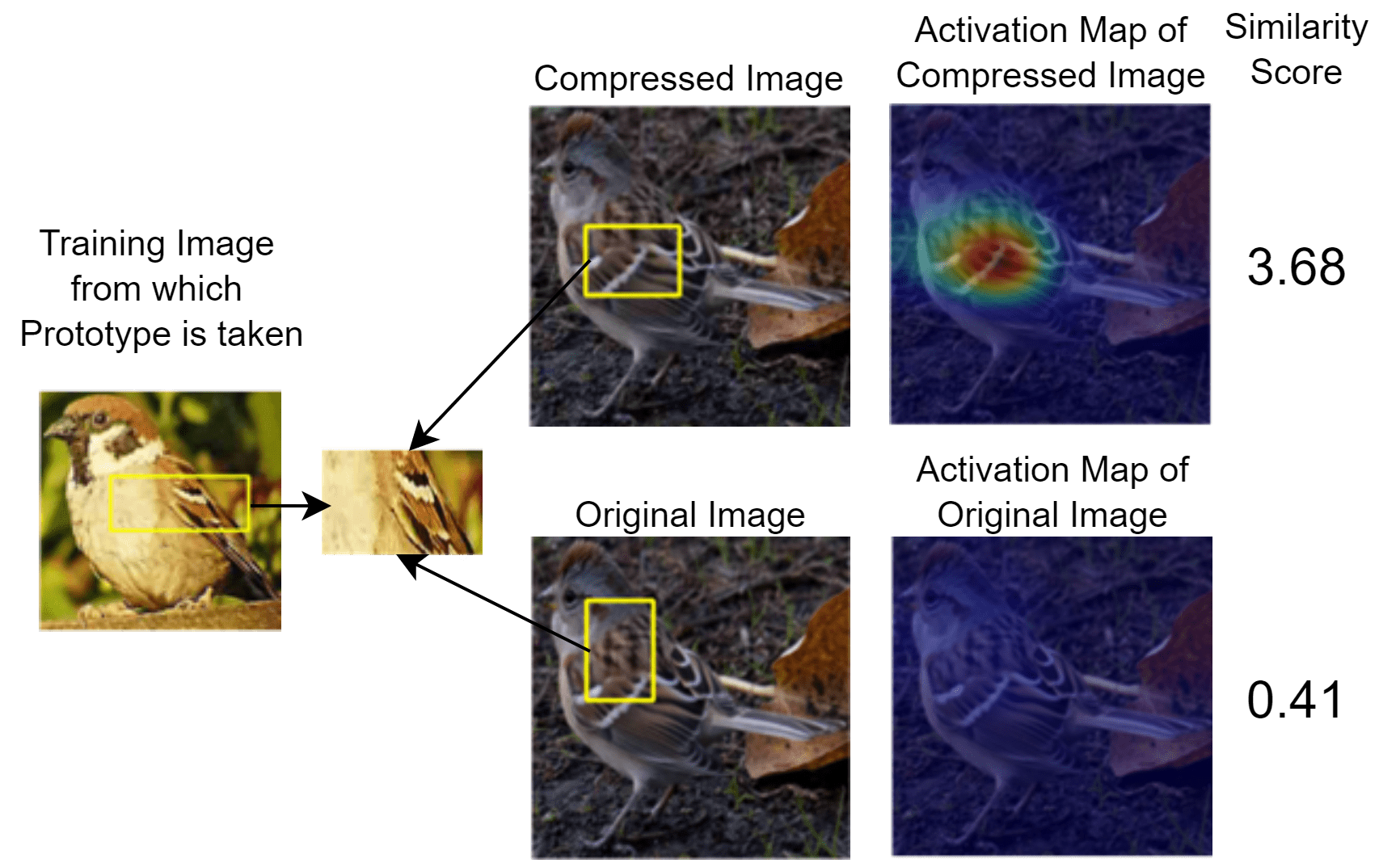}
  \caption{}
\end{subfigure}\hfill
\vspace{-1mm}
\caption{The JPEG experiment - ResNet-34}
\label{fig:results_jpeg_rn34}
\end{figure*}

\begin{figure*} [h]
\begin{subfigure}{0.32\textwidth}
  \centering
  \includegraphics[width=\linewidth]{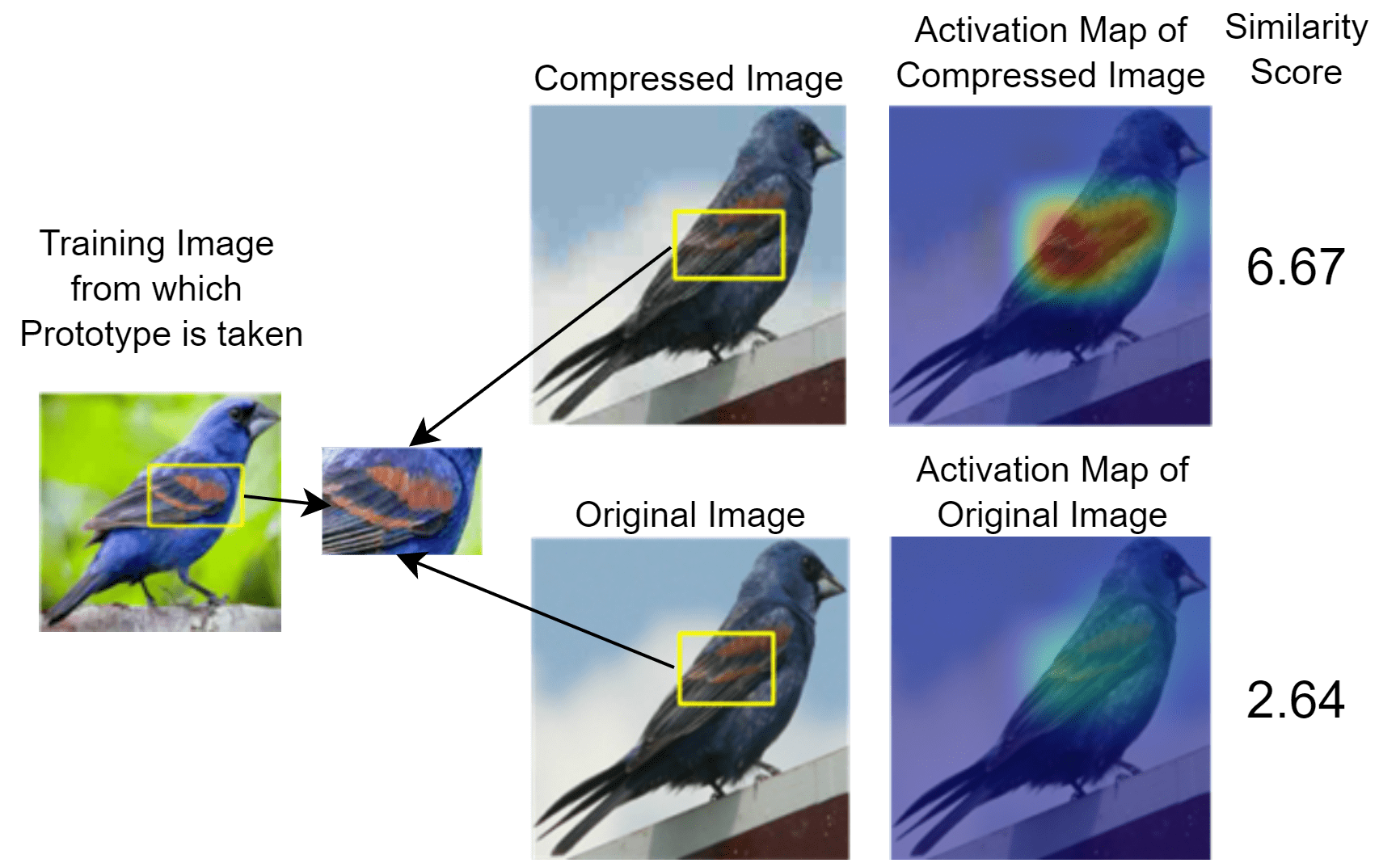}
  \caption{}
\end{subfigure}\hfill
\begin{subfigure}{0.32\textwidth}
  \centering
  \includegraphics[width=\linewidth]{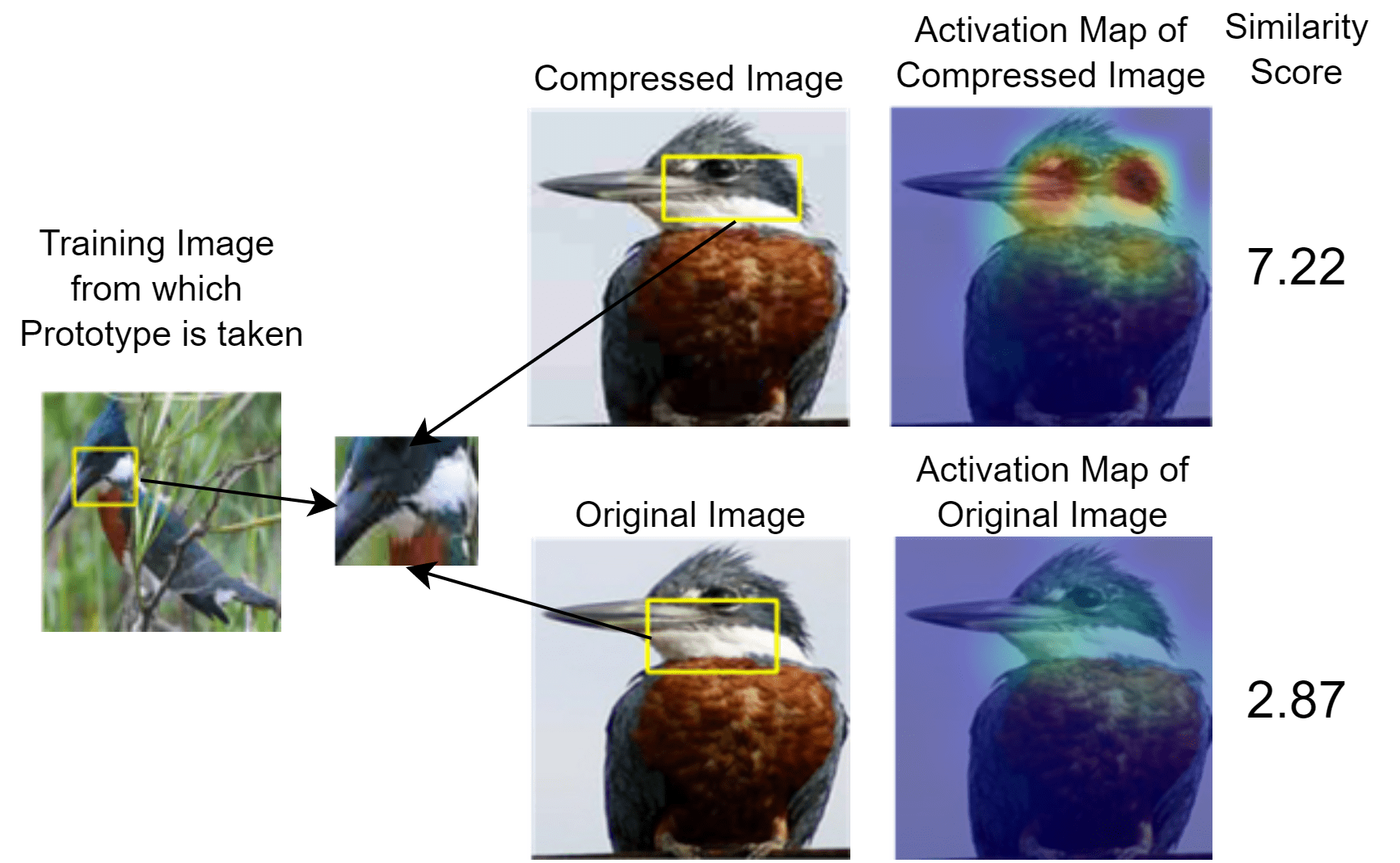}
  \caption{}
\end{subfigure}\hfill
\begin{subfigure}{0.32\textwidth}
  \centering
  \includegraphics[width=\linewidth]{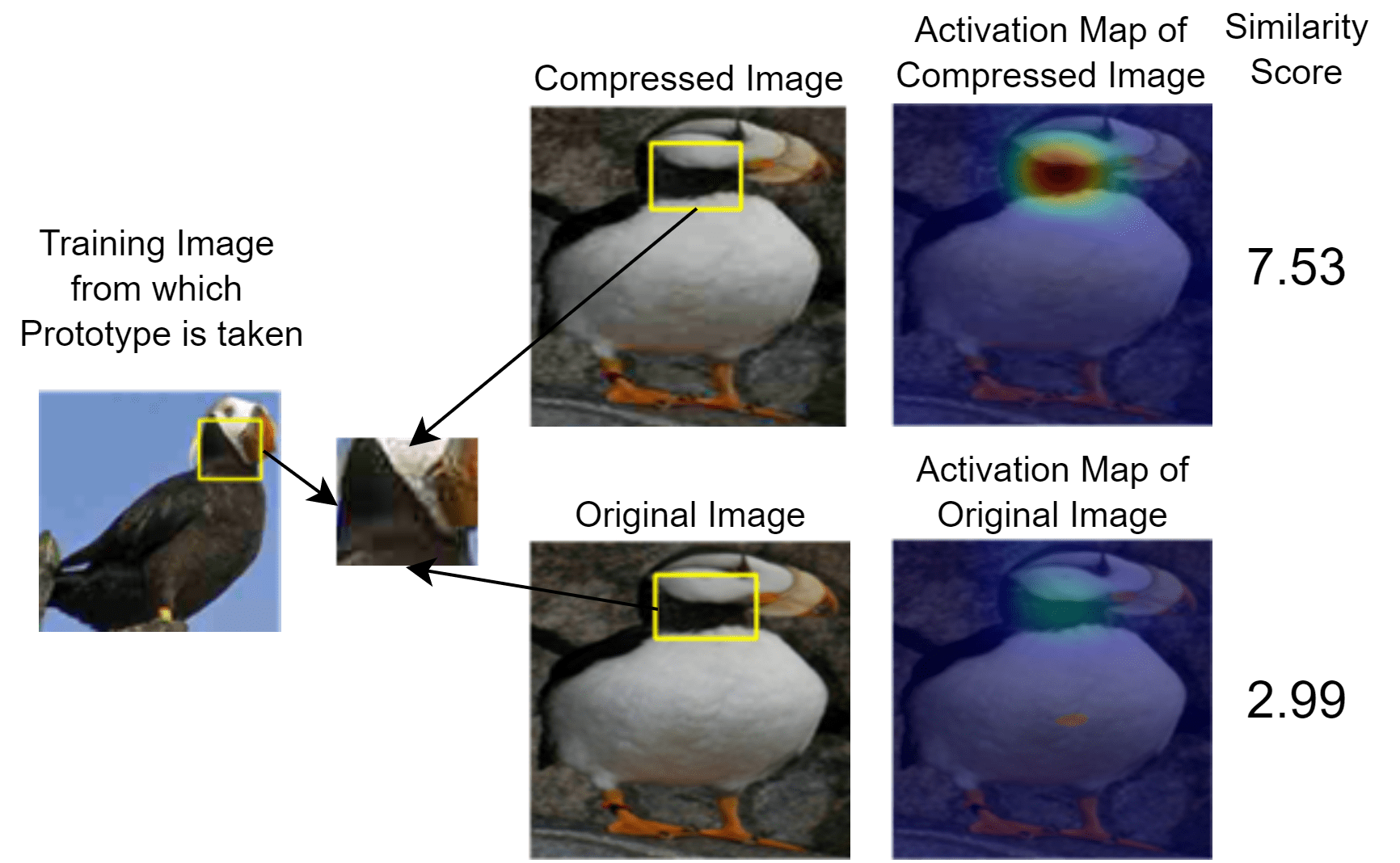}
  \caption{}
\end{subfigure}\hfill
\begin{subfigure}{0.32\textwidth}
  \centering
  \includegraphics[width=\linewidth]{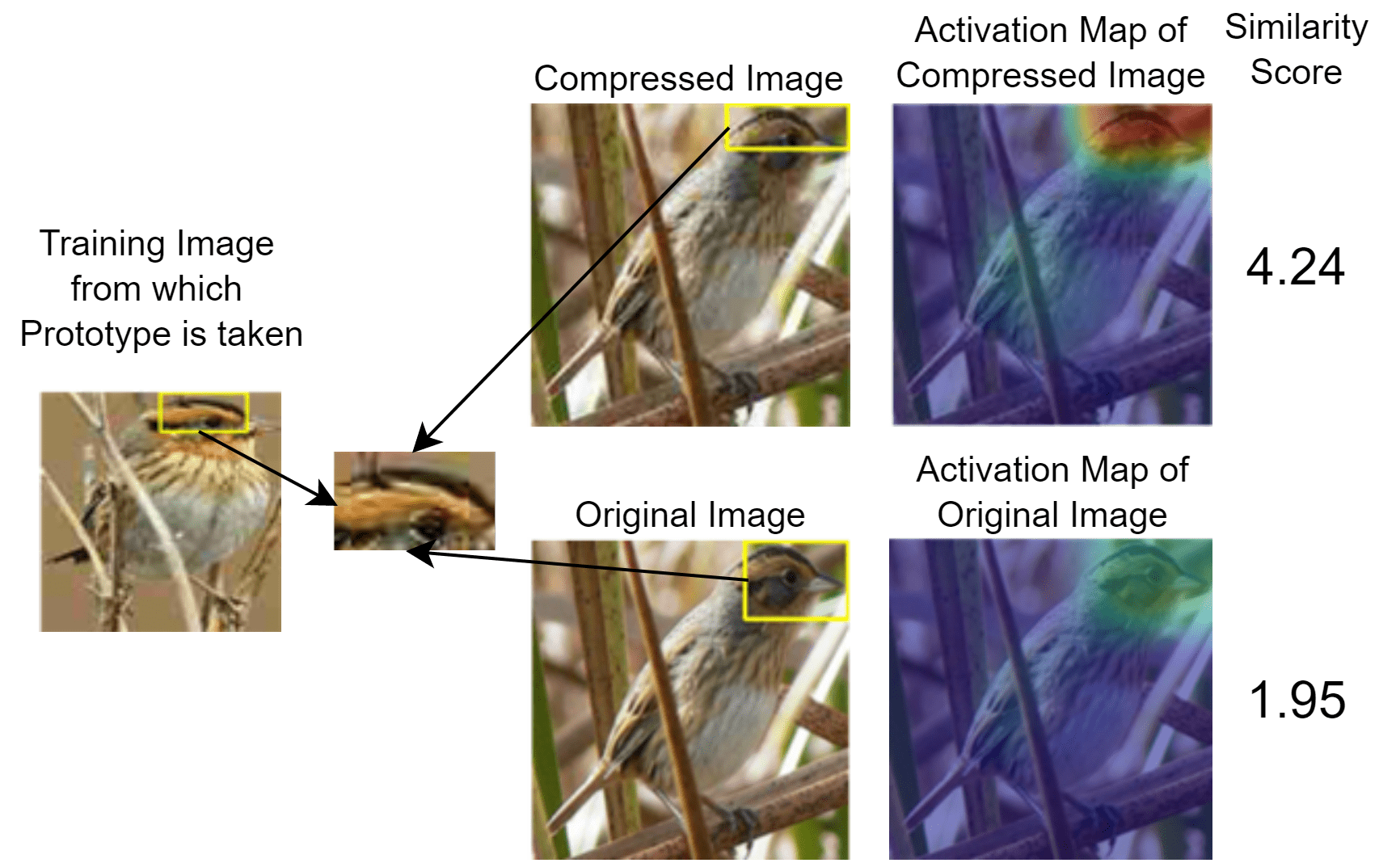}
  \caption{}
\end{subfigure}\hfill
\begin{subfigure}{0.32\textwidth}
  \centering
  \includegraphics[width=\linewidth]{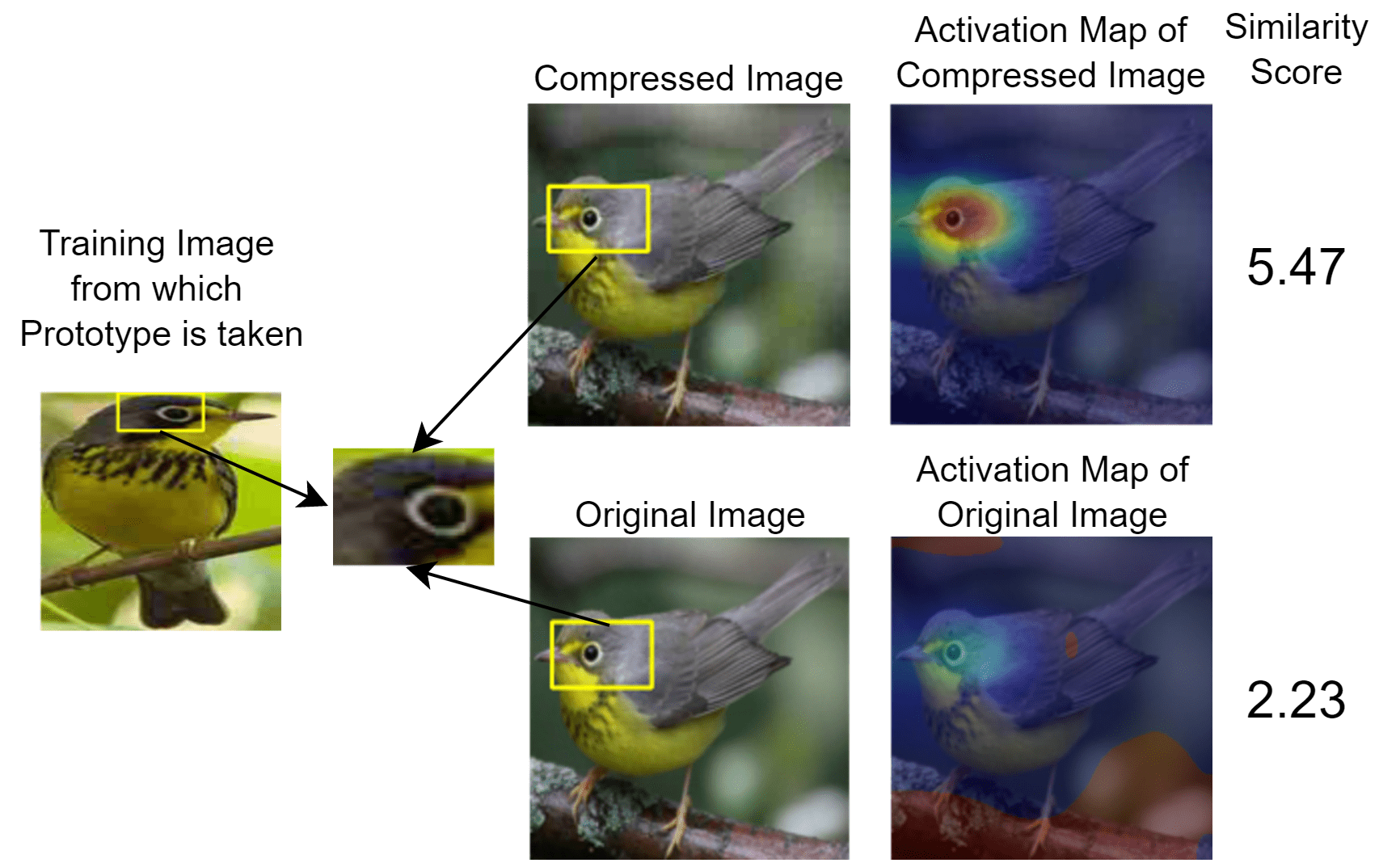}
  \caption{}
\end{subfigure}\hfill
\begin{subfigure}{0.32\textwidth}
  \centering
  \includegraphics[width=\linewidth]{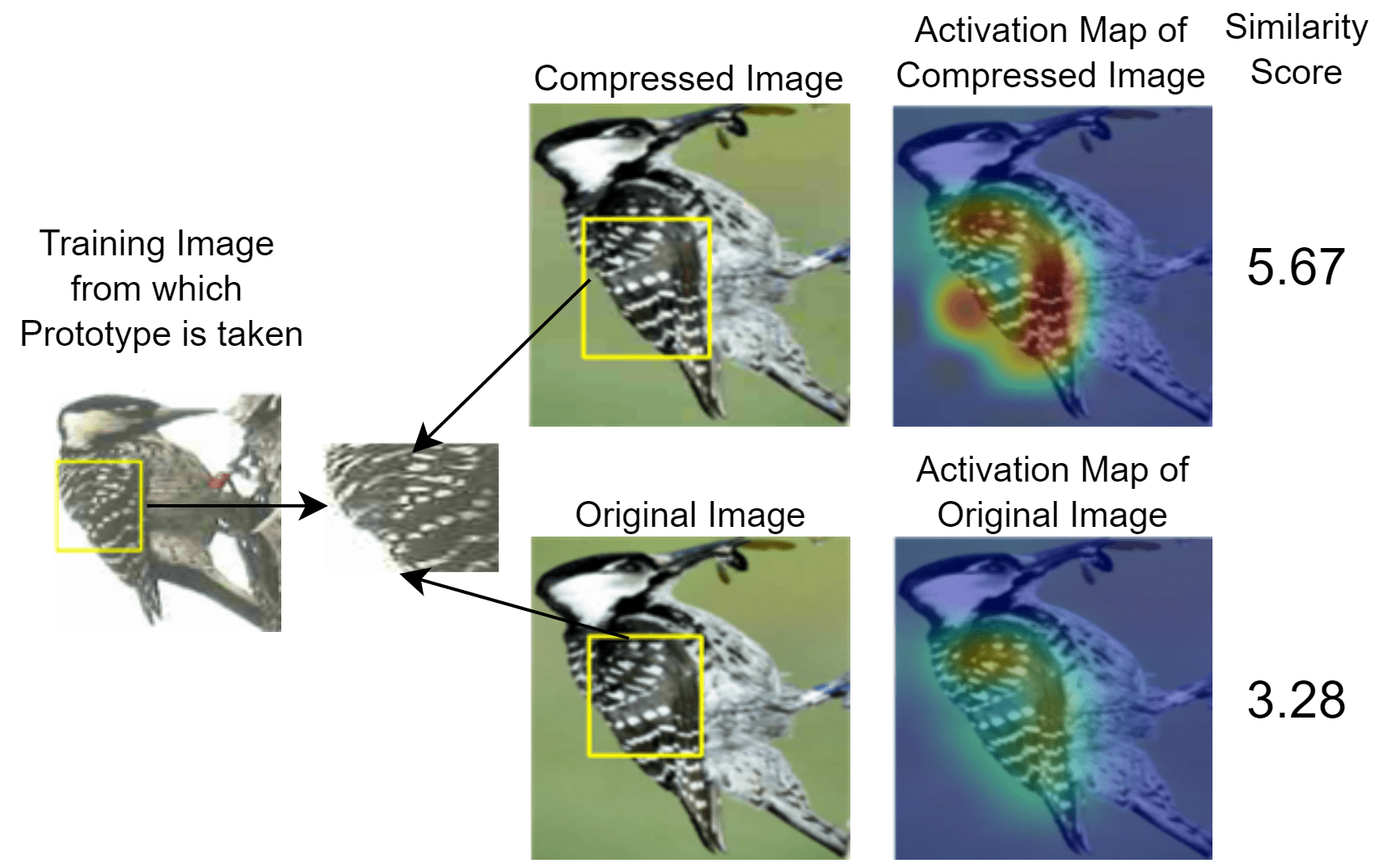}
  \caption{}
\end{subfigure}\hfill
\vspace{-1mm}
\caption{The JPEG experiment - VGG-19}
\label{fig:results_jpeg_vgg19}
\end{figure*}

In this subsection, we show further examples from the JPEG experiment which were successful. Figures~\ref{fig:results_jpeg_rn18},~\ref{fig:results_jpeg_rn34}~and~\ref{fig:results_jpeg_vgg19} present the results for ResNet-18, ResNet-34 and VGG-19 respectively. The results show that humans and ProtoPNets consider two image patches to be similar, but mostly for different reasons. This highlights a fundamental gap between our understanding of similarity and that of a ProtoPNet.

\FloatBarrier
\subsection{Histograms in JPEG Experiment} \label{app:jpeguniform}
\begin{figure*}[h]
  \centering
  \includegraphics[width=\linewidth]{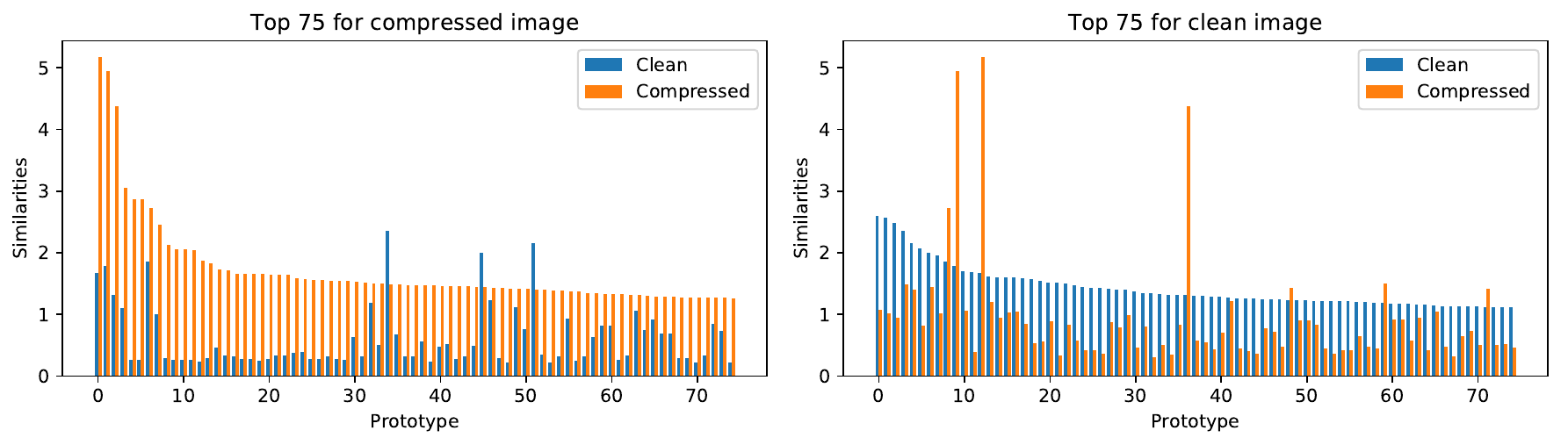}
  \caption{Highest similarities for the test image depicted in Figure~\ref{fig:results_jpeg_1} on ProtoPNet with ResNet-18 backbone from Section~\ref{sec:JPEG}.}
\label{fig:histo_resnet_18}
\end{figure*}

\begin{figure*}[h]
  \centering
  \includegraphics[width=\linewidth]{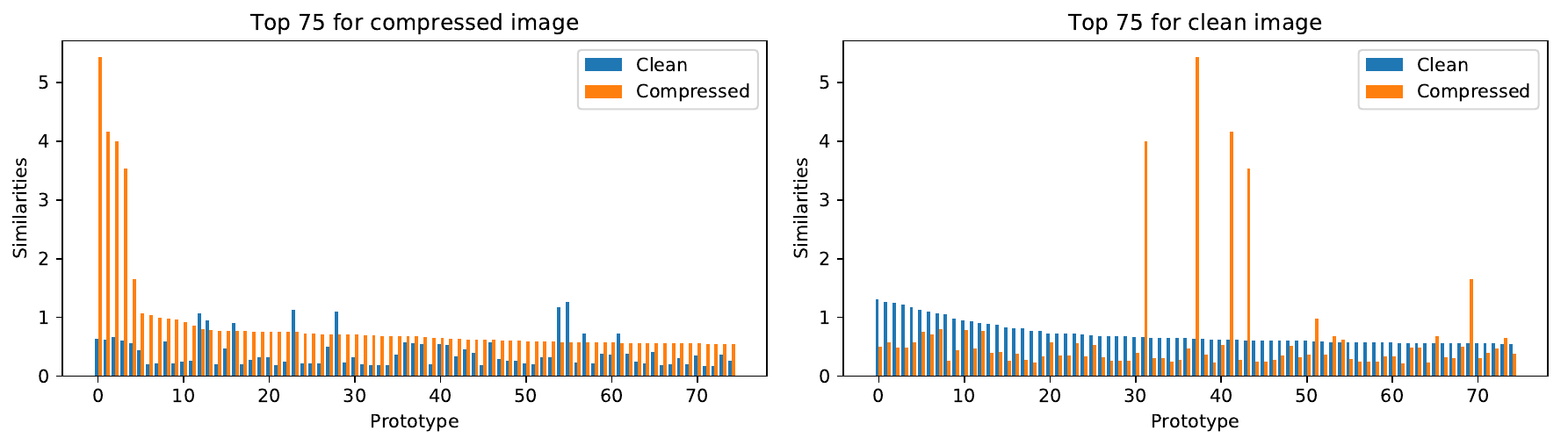}
  \caption{Highest similarities for the test image depicted in Figure~\ref{fig:results_jpeg_2} on ProtoPNet with ResNet-34 backbone from Section~\ref{sec:JPEG}.}
\label{fig:histo_resnet_34}
\end{figure*}

\begin{figure*}[h]
  \centering
  \includegraphics[width=\linewidth]{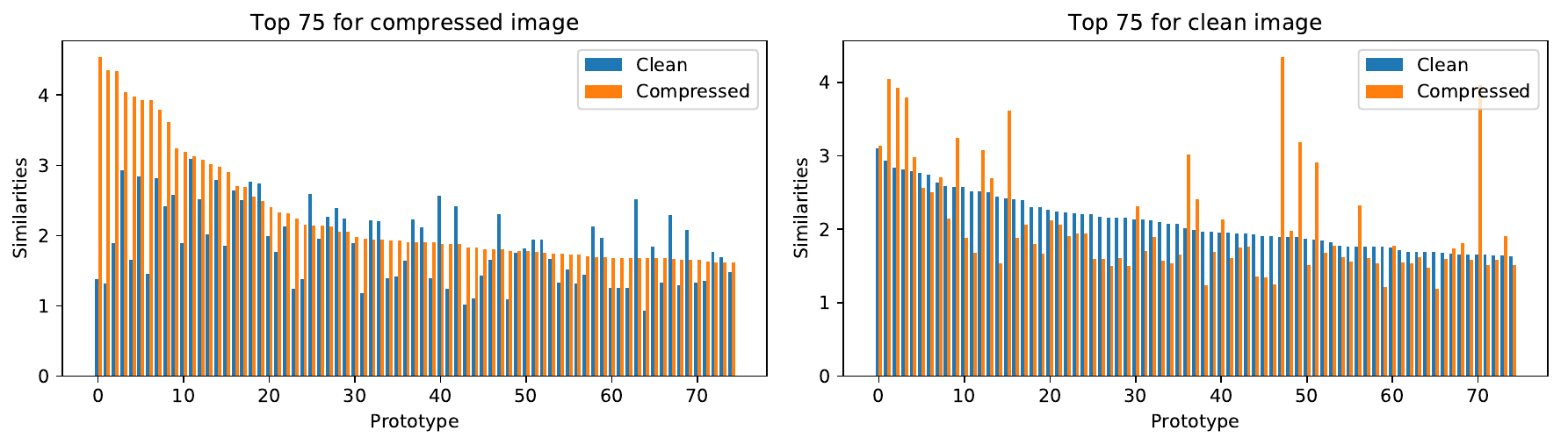}
  \caption{Highest similarities for the test image depicted in Figure~\ref{fig:results_jpeg_3} on ProtoPNet with VGG-19 backbone from Section~\ref{sec:JPEG}.}
\label{fig:histo_vgg19}
\end{figure*}

In Section~\ref{sec:JPEG}, we showed that if $\mathbf{p}_l$ is the prototype with the highest similarity reported by a ProtoPNet for some test image $\mathbf{\bar{x}}$ with JPEG compression artefacts, then the ProtoPNet might find a much lower similarity score for $\mathbf{p}_l$ if we remove the artefacts from $\mathbf{\bar{x}}$. This shows that ProtoPNets do not generally find compressed and clean versions of an image similar in the same manner as humans do. However, a ProtoPNet might uniformly scale down the similarity scores when switching from compressed to clean versions of an image. For example, if a ProtoPNet reports the highest similarity scores between a compressed test image $\mathbf{\bar{x}}$ and prototypes $\mathbf{p}_q$, $\mathbf{p}_r$, and $\mathbf{p}_s$ and does so again if we remove the noise then the ProtoPNet would consistently find the same prototypes to be most important. Which would be a very valuable characteristic of ProtoPNets. In Figures~\ref{fig:histo_resnet_18},~\ref{fig:histo_resnet_34},~and~\ref{fig:histo_vgg19}, we investigate this.
Each figure shows on the left the 75 similarity scores of the prototypes that are most strongly found by the ProtoPNet in an image $\mathbf{\bar{x}}$ (i.e. the JPEG compressed image) and overlaid the similarity scores of the respective prototypes but for image $\mathbf{x}$ (i.e. the clean version of $\mathbf{\bar{x}}$). The right parts of the plots show the other way round (i.e. the prototypes with the highest similarities to $\mathbf{x}$ are considered). The images and ProtoPNets we consider here are the same as in Figure~\ref{fig:results_jpeg}.

As can clearly be seen in the figures, the top few prototypes do not generally remain the prototypes with the highest similarity scores once the artefacts are removed and vice versa. This suggests that ProtoPNets do not consider the same prototypes to be the most important for compressed and clean versions of an image.

\FloatBarrier

\section{Performance of ProtoPNets used in our Experiments}\label{app:ppnetperformances}
Table~\ref{tab:results_acc} lists the test accuracy of the trained ProtoPNets used in the Head on Stomach experiment. We obtain similar performance as reported in \citet{chen2018looks} for all networks except VGG-19. We suspect that they use VGG-19 with batch normalization instead of vanilla VGG-19, while we use vanilla VGG-19 for our experiments. Notably, we achieve a test accuracy of 79.24\% when using VGG-19 with batch normalization. Lastly, the test accuracies for the models used in the JPEG experiment are shown in Table~\ref{tab:results_jpeg_acc}.

\begin{minipage}{\textwidth}
    \begin{minipage}[t]{0.49\textwidth}
        \centering
            \begin{tabular}{l|c|c}
                \hline \text { Backbone } & \text { Chen et al } & \text { Ours } \\
                \hline \text { ResNet-18 } & - & 75.75 \\
                \text { ResNet-34 } & $79.2 \pm 0.1$ & 79.58 \\
                \text { VGG-19 } & $78.0 \pm 0.2$ & 76.39 \\
                \hline   
            \end{tabular}
            \captionof{table}{Test accuracies of the trained ProtoPNets used in Head on Stomach experiment. We add the corresponding numbers from \citet{chen2018looks} for comparison.}
            \label{tab:results_acc}
      \end{minipage}
  \hspace*{2mm}
  \begin{minipage}[t]{0.49\textwidth}
    \centering
        \begin{tabular}{l|c|c}
        \hline \text { Backbone } & \text { Altered test set } & \text { Original test set } \\
        \hline \text { ResNet-18 } &80.19  &44.75\\
        \text { ResNet-34 } &82.14  &46.12\\
        \text { VGG-19 } &81.08  &49.60\\
        \hline
        \end{tabular}
        \captionof{table}{Test accuracies of the ProtoPNets used in JPEG experiment. Specifically, for the altered dataset we use the CUB-200-2011 test set and apply the same compression scheme as during training. For the original test set we simply take the CUB-200-2011 test set.}
        \label{tab:results_jpeg_acc}
  \end{minipage}
\end{minipage}

\section{Countermeasures to Improve Robustness}\label{sec:countermeasures}
\subsection{Remedy Head on Stomach Experiment} \label{app:remadvtraining}
In Section~\ref{sec:head_on_stomach}, we showed that the interpretability mechanism of ProtoPNet can be abused to identify prototypes at arbitrary locations in the test image. Since the experiment utilises ideas from adversarial attacks~\cite{szegedy2014, goodfellow2015}, it is natural to test if remedies from that domain work here as well. Specifically, we train ProtoPNets with ResNet-18 and ResNet-34 backbones via fast adversarial training \cite{wong2020fast}. Note however that it is not clear a priori if this helps against the Head on Stomach problem. This since adversarial training was developed to help classify images correctly even if you add noise to an image. In our case, we do not have the problem of misclassification (see Section~\ref{sec:head_on_stomach}). Instead, we observe the issue that the latent representations admit a change in similarity scores with slight variations in the input space.

We perform adversarial training via fast gradient sign method (FGSM)~\cite{goodfellow2015} with random initialization and set step size $\alpha = 10/255$, $\ell_\infty$ radius $\varepsilon = 8/255$, following~\citet{wong2020fast}. We train the models for $10$ epochs and use the same training setup as mentioned in Appendix~\ref{app:implementation-details}. The performance of our trained models is listed in Table~\ref{tab:susceptible}. The adversarial accuracy is evaluated using PGD with step size $2/255$, $\varepsilon = 8/255$ and $10$ iterations. 

Lastly, we devise a measure to quantify the susceptibility of ProtoPNet's interpretations to the Head on Stomach problem. For a ProtoPNet and test image $\mathbf{x}$, we consider the top $k$ ($=5$) prototypes with the highest similarity (denoted $\{\mathbf{p}_{l_i}\}_{i=1}^{k}$) among all prototypes. Suppose the prototype $\mathbf{p}_{l_i}$ is recognized at the patch $S$ of the latent representation of image $\mathbf{x}$ and $A$ is the set of all patches in the latent representation of $\mathbf{x}$. We then try to perturb $\mathbf{x}$ with the goal that the ProtoPNet finds any arbitrary patch $S_{\text{noisy}} \in A \setminus S$ highly similar to $\mathbf{p}_{l_i}$ by maximizing the objective of Equation~\ref{eq:head-on-stomach-objective}. Precisely, we solve the optimization problem stated in Equation~\ref{eq:head-on-stomach-optimization} using PGD with step size $2/255$, $\varepsilon = 8/255$ and apply $40$ iterations. Finally, if the similarity of $\mathbf{p}_{l_i}$ with $S_{\text{noisy}}$ is greater than that with $S$, we regard the experiment as being successful. We do this for each of the top $k$ prototypes and consider the experiment as a success for image $\mathbf{x}$ if it succeeds for any of the $k$ prototypes. We repeat this procedure for a random subset of 200 test images that are correctly classified by the ProtoPNet under consideration and compute the success rate of our algorithm. 

In Table~\ref{tab:susceptible}, we also list the results of our evaluation. As we can see, the adversarial training reduces the success rate of our algorithm by more than half but does not alleviate the problem entirely. Additionally, the robustness to our algorithm comes at a significant drop in accuracy on the test set (in line with adversarial training methods \cite{tsipras2018robustness, trades2019}).
\begin{table} [h] 
    \centering
    \begin{tabular}{l|ccc|ccc}
        \hline \multirow{2}{*}{\text { Backbone }} & \multicolumn{3}{c|}{\text { Standard Training}} & \multicolumn{3}{c}{\text { Adversarial Training }} \\
        & Clean Acc. & Adverarial Acc. & Success Rate & Clean Acc. & Adverarial Acc. & Success Rate \\
        \hline \text { ResNet-18 } & 75.8 & 0.0 & 99.5 & 56.3 & 15.8 & 34.0 \\
        \text { ResNet-34 } & 79.6 & 0.0 & 93.0 & 58.8 & 24.2 & 27.0 \\
        \hline   
    \end{tabular}
    \caption{Performance of ProtoPNets trained via standard and adversarial training respectively on the CUB-200-2011 test set. Clean and adversarial accuracy are evaluated on the entire test set while the success rate is computed on a random subset of 200 correctly classified test images.}
    \label{tab:susceptible}
\end{table}


\subsection{Remedy JPEG Experiment} \label{app:remcompression}
In Section~\ref{sec:JPEG}, we showed that the interpretations provided by ProtoPNets can behave in a manner that we humans cannot comprehend if JPEG artefacts are introduced in the training data. In this section, we investigate if the problem can be alleviated by augmenting the training data with JPEG compressed images. Specifically, we train ProtoPNets for the same backbones as in Section~\ref{sec:JPEG} (i.e. ResNet-18, ResNet-34, VGG-19) but add JPEG compression to the augmentation pipeline. To be exact, every image in a mini-batch has a 50\% chance of being JPEG compressed with 20\% compression quality during training. All other augmentations and training parameters are as in \citet{chen2018looks} for the CUB-200-2011 dataset. The new test accuracies for the models trained with JPEG augmentation are shown in Table~\ref{tab:results_jpeg_acc_rem}.

\begin{figure*} [h]
  \centering
  \includegraphics[width=\linewidth]{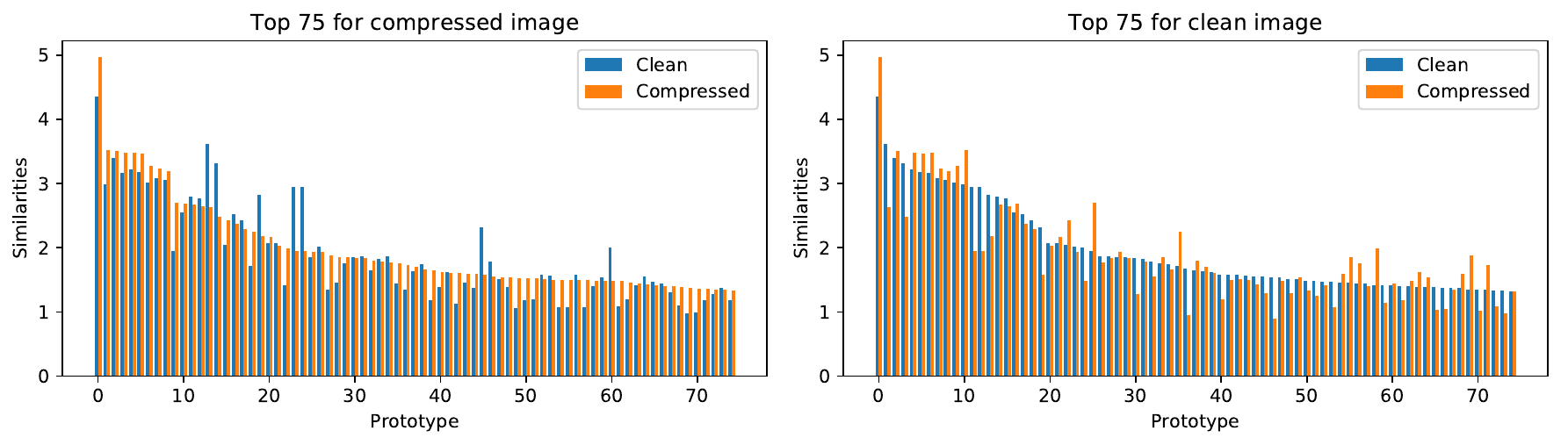}
\caption{Highest similarities for the test image depicted in Figure~\ref{fig:results_jpeg_1} on ProtoPNet with ResNet-18 backbone that was trained with JPEG compression as an augmentation step.}
\label{fig:histo_resnet_18_robust}
\end{figure*}

\begin{figure*} [h]
  \centering
  \includegraphics[width=\linewidth]{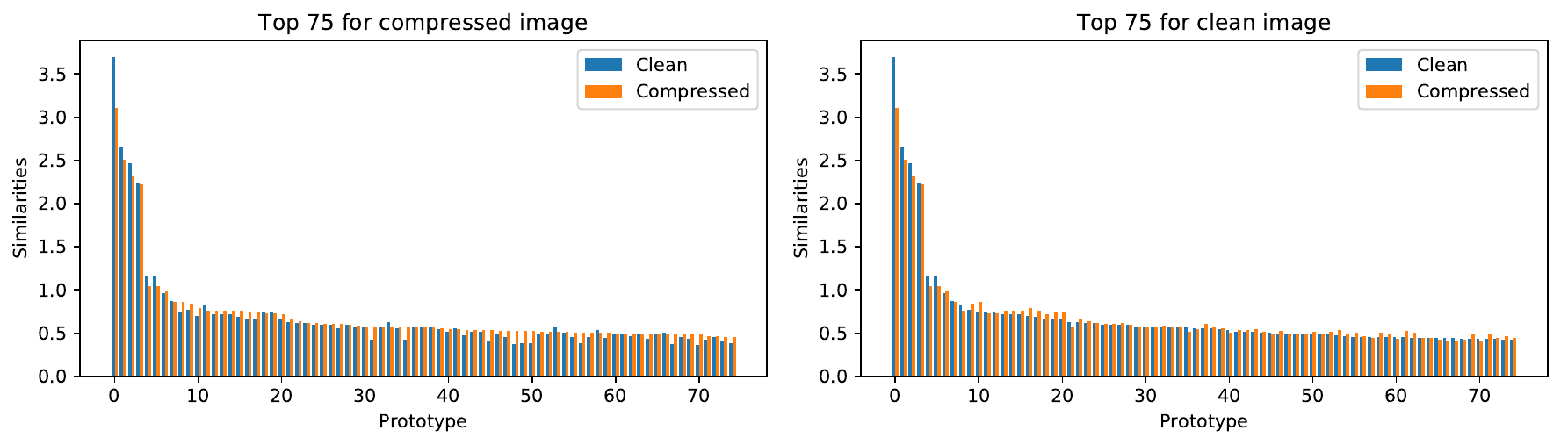}\\
\caption{Highest similarities for the test image depicted in Figure~\ref{fig:results_jpeg_2} on ProtoPNet with ResNet-34 backbone that was trained with JPEG compression as an augmentation step.}
\label{fig:histo_resnet_34_robust}
\end{figure*}

\begin{figure*} [h]
  \centering
  \includegraphics[width=\linewidth]{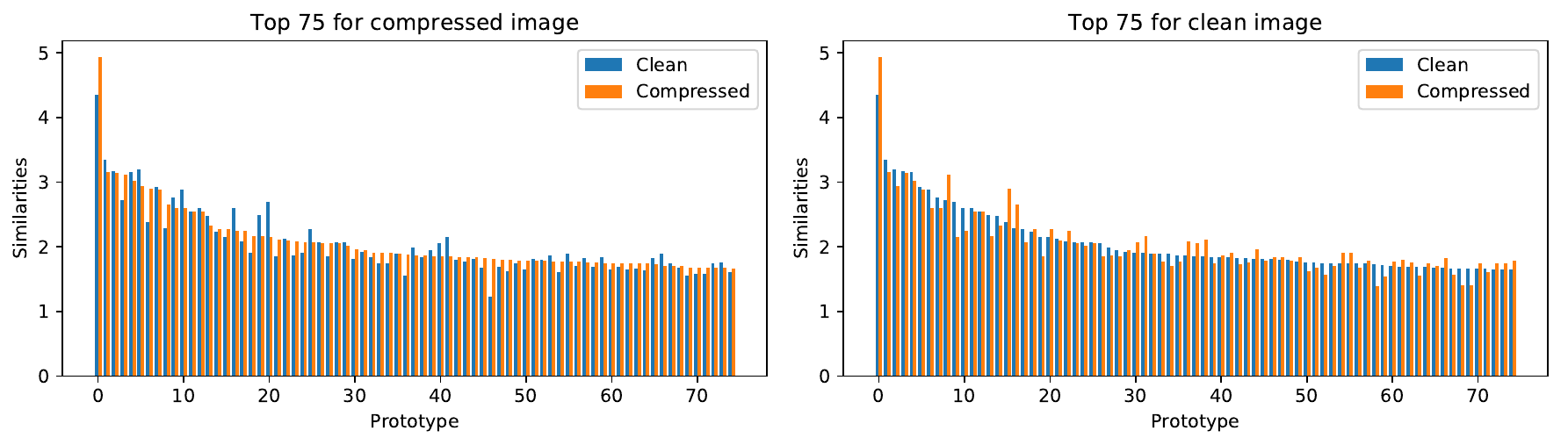}
\caption{Highest similarities for the test image depicted in Figure~\ref{fig:results_jpeg_3} on ProtoPNet with VGG-19 backbone that was trained with JPEG compression as an augmentation step.}
\label{fig:histo_vgg19_robust}
\end{figure*}

Figures~\ref{fig:histo_resnet_18_robust},~\ref{fig:histo_resnet_34_robust},~and~\ref{fig:histo_vgg19_robust} depict the same visualisation of test images as shown in Figures~\ref{fig:histo_resnet_18},~\ref{fig:histo_resnet_34},~and~\ref{fig:histo_vgg19} respectively but for the newly trained ProtoPNets. As can be seen, the similarities of the top prototypes behave much less erratic than in Appendix~\ref{app:jpeguniform}. This suggests that the augmentation helps to alleviate the problem encountered in the JPEG experiment where the presence or absence of compression artefacts affected similarities in a way that humans find counter-intuitive. Yet, the problem is not completely resolved by the added augmentation step.

\begin{table}[h] 
    \centering
    \begin{tabular}{l|c|c}
    \hline \text { Backbone } & \text { Altered test set } & \text { Original test set } \\
    \hline \text { ResNet-18 } &74.70  &76.32\\
    \text { ResNet-34 } &76.58  &77.70\\
    \text { VGG-19 } &74.59  &76.30\\
    \hline
    \end{tabular}
    \caption{Test accuracies of the ProtoPNets used in JPEG experiment with random JPEG augmentation during training. The altered dataset and the original dataset follow the same scheme as in Table \ref{tab:results_jpeg_acc}.}
    \label{tab:results_jpeg_acc_rem}
\end{table}

\FloatBarrier